\documentclass[manuscript,screen]{acmart}

\AtBeginDocument{%
  }

\setcopyright{acmlicensed}
\copyrightyear{2018}
\acmYear{2018}
\acmDOI{XXXXXXX.XXXXXXX}

\acmConference[Conference acronym 'XX]{Make sure to enter the correct
  conference title from your rights confirmation emai}{June 03--05,
  2018}{Woodstock, NY}
\acmISBN{978-1-4503-XXXX-X/18/06}
\usepackage[edges]{forest}
\usepackage{amsmath}
\usepackage{url}
\usepackage{graphicx}
\usepackage{enumitem}
\usepackage{subfig}
\usepackage{amsmath}
\usepackage{multirow}
\usepackage{tabularx, booktabs}
\usepackage{pgfplots}
\usepackage{listings}
\usepackage{bbm}
\usepackage{adjustbox}
\usepackage{wrapfig}
\usepackage{xcolor}
\usepackage{natbib}
\forestset{
  level 0/.style={
    fill=gray!20,
    draw=gray!70,
    thick,
    text width=17em,
    align=left
  },
  level 1/.style={
    fill=blue!15,
    draw=blue!60,
    thick,
    text width=14em,
    align=left
  },
  level 2/.style={
    fill=teal!15,
    draw=teal!60,
    thick,
    text width=13em,
    align=left
  },
  level 3/.style={
    fill=orange!15,
    draw=orange!60,
    thick,
    text width=12em,
    align=left
  },
  default preamble={
    for tree={
      grow=east,
      reversed=true,
      anchor=base west,
      parent anchor=east,
      child anchor=west,
      font=\large,
      rectangle,
      rounded corners,
      minimum width=4em,
      edge={darkgray, line width=1pt},
      s sep=3pt,
      inner xsep=2pt,
      inner ysep=3pt,
      line width=0.8pt,
      align=left
    }
  },
}

\definecolor{beautifulblue}{RGB}{219,227,242}
\definecolor{beautifulred}{RGB}{248,225,216}
\definecolor{mydarkblue}{rgb}{0,0.08,0.45}

\usepackage{float}

\usepackage{xcolor}
\definecolor{liteblue}{RGB}{120,170,240}
\definecolor{liteyellow}{RGB}{255,190,100}
\definecolor{litered}{RGB}{250,160,160}

\begin{document}

%%
%% The "title" command has an optional parameter,
%% allowing the author to define a "short title" to be used in page headers.
\title{Usable XAI: 10 Strategies Towards Exploiting Explainability in the LLM Era}

%%
%% The "author" command and its associated commands are used to define
%% the authors and their affiliations.
%% Of note is the shared affiliation of the first two authors, and the
%% "authornote" and "authornotemark" commands
%% used to denote shared contribution to the research.
%\author{Anonymous Author}
%\authornote{The authors contributed equally to this research.}
%\email{xuansheng.wu@uga.edu}
%\orcid{1234-5678-9012}
%\affiliation{%
%  \institution{Anonymous Institution}
%}

\author{Xuansheng Wu}
\authornote{The authors contributed equally to this research.}
\email{xuansheng.wu@uga.edu}
%\orcid{1234-5678-9012}
\affiliation{%
  \institution{University of Georgia}
  \city{Athens}
  \state{Georgia}
  \country{USA}
}

\author{Haiyan Zhao}
\authornotemark[1]
\email{hz54@njit.edu}
%\orcid{1234-5678-9012}
\affiliation{%
  \institution{New Jersey Institute of Technology}
  \city{Newark}
  \state{New Jersey}
  \country{USA}
}

\author{Yaochen Zhu}
\authornotemark[1]
\email{uqp4qh@virginia.edu}
%\orcid{1234-5678-9012}
\affiliation{%
  \institution{University of Virginia}
  \city{Charlottesville}
  \state{Virginia}
  \country{USA}
}

\author{Yucheng Shi}
\authornotemark[1]
\email{yucheng.shi@uga.edu}
%\orcid{1234-5678-9012}
\affiliation{%
  \institution{University of Georgia}
  \city{Athens}
  \state{Georgia}
  \country{USA}
}

\author{Fan Yang}
\email{yangfan@wfu.edu}
%\orcid{1234-5678-9012}
\affiliation{%
  \institution{Wake Forest University}
  \city{Winston-Salem}
  \state{North Carolina}
  \country{USA}
}

\author{Lijie Hu}
\email{lijie.hu@kaust.edu.sa}
%\orcid{1234-5678-9012}
\affiliation{%
  \institution{King Abdullah University of Science and Technology}
  \country{Saudi Arabia}
}

\author{Tianming Liu}
\email{tliu@uga.edu}
%\orcid{1234-5678-9012}
\affiliation{%
  \institution{University of Georgia}
  \city{Athens}
  \state{Georgia}
  \country{USA}
}

\author{Xiaoming Zhai}
\email{xiaoming.zhai@uga.edu}
%\orcid{1234-5678-9012}
\affiliation{%
  \institution{University of Georgia}
  \city{Athens}
  \state{Georgia}
  \country{USA}
}

\author{Wenlin Yao}
\email{ywenlin@amazon.com}
%\orcid{1234-5678-9012}
\affiliation{%
  \institution{Amazon}
  \city{Seattle}
  \state{Washington}
  \country{USA}
}

\author{Jundong Li}
\email{jundong@virginia.edu}
%\orcid{1234-5678-9012}
\affiliation{%
  \institution{University of Virginia}
  \city{Charlottesville}
  \state{Virginia}
  \country{USA}
}

\author{Mengnan Du}
\email{mengnan.du@njit.edu}
\affiliation{%
  \institution{New Jersey Institute of Technology}
  \city{Athens}
  \state{Georgia}
  \country{USA}
}
\author{Ninghao Liu}
\email{ninghao.liu@uga.edu}
%\orcid{1234-5678-9012}
\affiliation{%
  \institution{University of Georgia}
  \city{Athens}
  \state{Georgia}
  \country{USA}
}

%%
%% By default, the full list of authors will be used in the page
%% headers. Often, this list is too long, and will overlap
%% other information printed in the page headers. This command allows
%% the author to define a more concise list
%% of authors' names for this purpose.
\renewcommand{\shortauthors}{Wu et al., 2025}

%%
%% The abstract is a short summary of the work to be presented in the
%% article.
\begin{abstract}
Explainable AI (XAI) refers to techniques that provide human-understandable insights into the workings of AI models. 
Recently, the focus of XAI is being extended toward explaining Large Language Models (LLMs). 
This extension calls for a significant transformation in the XAI methodologies for two reasons. First, many existing XAI methods cannot be directly applied to LLMs due to their complexity and advanced capabilities. Second, as LLMs are increasingly deployed in diverse applications, the role of XAI shifts from merely opening the ``black box'' to actively enhancing the productivity and applicability of LLMs in real-world settings.
Meanwhile, the conversation and generation abilities of LLMs can reciprocally enhance XAI.
Therefore, in this paper, we introduce Usable XAI in the context of LLMs by analyzing (1) how XAI can explain and improve LLM-based AI systems and (2) how XAI techniques can be improved by using LLMs. We introduce 10 strategies, introducing the key techniques for each and discussing their associated challenges. We also provide case studies to demonstrate how to obtain and leverage explanations.
The code used in this paper can be found at: \url{https://github.com/JacksonWuxs/UsableXAI_LLM}.
\end{abstract}

%%
%% The code below is generated by the tool at http://dl.acm.org/ccs.cfm.
%% Please copy and paste the code instead of the example below.
%%


%%
%% Keywords. The author(s) should pick words that accurately describe
%% the work being presented. Separate the keywords with commas.

% \received{20 February 2007}
% \received[revised]{12 March 2009}
% \received[accepted]{5 June 2009}

%%
%% This command processes the author and affiliation and title
%% information and builds the first part of the formatted document.
\maketitle

% \clearpage

\section{Introduction}
Explainability holds great promise for understanding machine learning models and providing directions for improvement. 
In practice, users have high expectations for model explainability: 
\begin{center}
    \vspace{-3pt}
    \textit{\textbf{1. Through explanation, can we know if a model works properly?} }
    \vspace{-2pt}
\end{center}
\begin{center}
    \vspace{-2pt}
    \textit{\textbf{2. Does explainability tell us how to develop better models?} }
    \vspace{-3pt}
\end{center}
First, explanations are expected to illuminate whether a model operates as humans expect. For example, does the model leverage reliable evidence and domain knowledge in its decision making? Does the model contain bias and discrimination? Does the model show any vulnerabilities to potential attacks? Will the model output harmful information?
Second, in recognition of model imperfections, we aspire for explainability to inform the development of better models. For example, how to adjust the behaviors of a model if we find that it uses unreliable or unreasonable features in making predictions? Can we improve the performance of a model by aligning its behavior with human preferences?

Therefore, the question arises: \textbf{Have these expectations been met?} 
Since the rise of deep learning, the body of literature on Explainable AI (XAI) has expanded rapidly to improve model transparency~\citep{du2019techniques,murdoch2019definitions,tjoa2020survey,dovsilovic2018explainable,rudin2022interpretable}, encompassing a wide array of methods customized for different data modalities, including visual~\citep{zhang2018visual}, textual~\citep{danilevsky2020survey}, graph~\citep{yuan2022explainability}, and time-series data~\citep{zhao2023interpretation}. 
% Some literature delves into specific techniques, such as attention methods, generalized additive models, and causal models.
In addition, some offer reviews of general principles and initiate discussions on evaluating the faithfulness of explanations~\citep{yang2019evaluating,doshi2017towards}.
\textbf{Despite the progress, the last mile of XAI -- making use of explanations -- has not received the attention it merits.}
In many cases, we seem to be satisfied with just acquiring explanations and their associated visualizations, sometimes followed by qualitative discussions of the model's strengths and failure cases. However, how to quantify model properties (e.g., fairness, security, rationality) and improve models by utilizing explanations remains a difficult task.

% The challenges in making use of explainability are twofold.
% First, there is an inherent conflict between AI automation and human engagement in XAI.
% Humans need to scrutinize explanation to identify if any vulnerabilities exist in the model, or define explainability that the model should follow. However, the requirement for human oversight introduces substantial costs, posing challenges to the scalability and practical implementation of model debugging and improvement in AI workflows.
% Second, many of the current approaches view explainability as a purely technical matter, ignoring the needs of practitioners and non-technical stakeholders.
% Existing XAI methods are developed mainly as statistical and mathematical tools. However, there exists a noticeable disparity between the objectives of these tools and the expectations of practitioners across various application domains~\citep{malizia2023current}. An explanation that satisfies a technical audience might offer little value to a non-technical audience.

While the opacity issues have not yet been fully resolved for traditional deep models (e.g., multi-layer perceptrons, convolutional and recurrent neural networks), the recent advancements of Large Language Models (LLMs)~\citep{brown2020language,achiam2023gpt,touvron2023llama2,chiang2023vicuna} appear to have exacerbated the challenge we are facing.
Firstly, LLMs typically possess a significantly larger model size and a greater number of parameters. This increased model complexity intensifies the difficulty of explaining their inner workings.
Second, unlike traditional ML models that primarily focus on low-level pattern recognition tasks such as classification and parsing, LLMs can handle more complex tasks~\cite{zhao2023survey} such as generation, reasoning, and question answering. Understanding the exclusive abilities of LLMs presents novel challenges for XAI techniques.
Considering the transformative impact of LLMs across various applications, ensuring the explainability and ethical use of LLMs has become an imminent and pressing need.
Meanwhile, the emerging capabilities of LLMs also present new opportunities for XAI research. Their human-like communication and commonsense reasoning skills offer prospects for achieving explainability in ways that could potentially augment or replace human involvement.

\begin{figure}[t]
\begin{center}\includegraphics[width=1.0\textwidth]{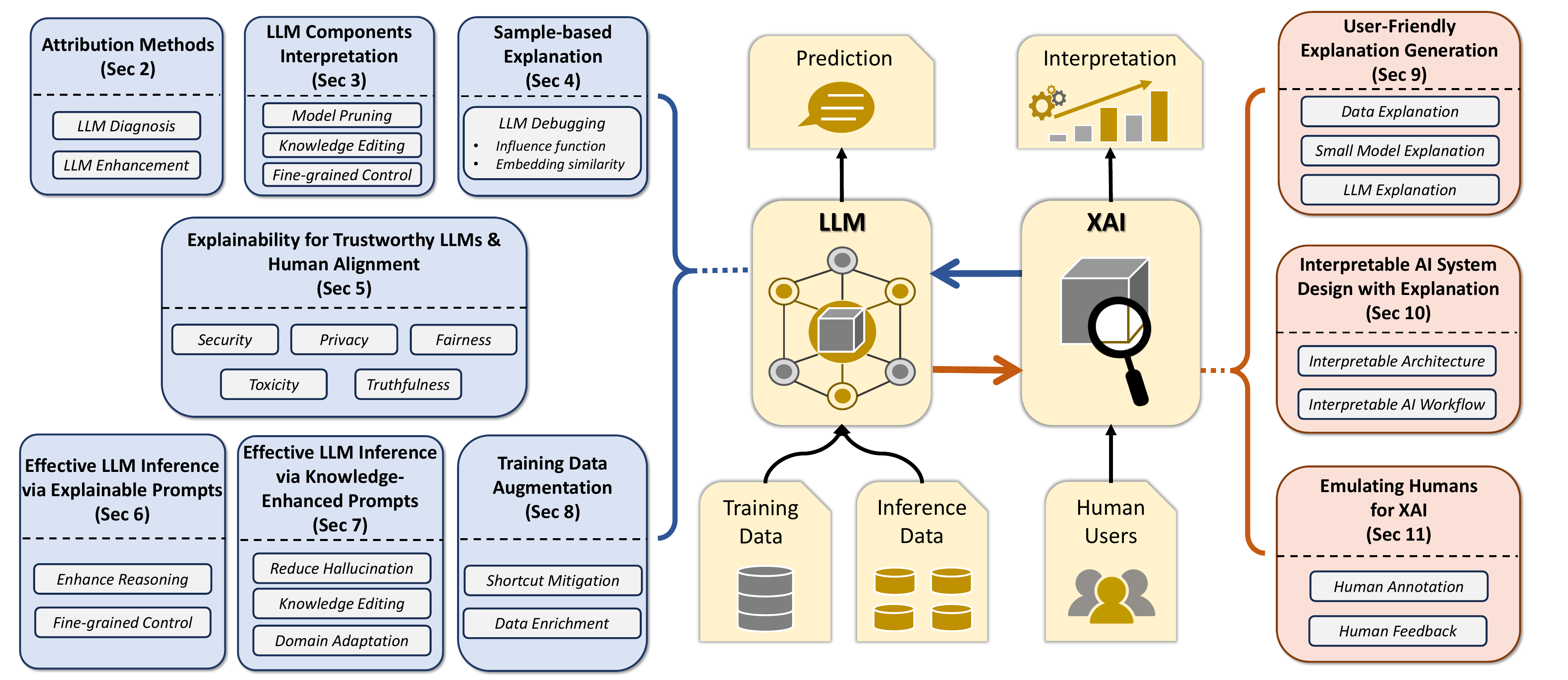}
\end{center}
  \vspace{-.2cm}
  \caption{The contributions and outline of this paper. We define Usable XAI in the context of LLMs with \colorbox{beautifulblue}{seven strategies} of enhancing LLMs with XAI, and \colorbox{beautifulred}{three strategies} of enhancing XAI with LLMs.}\label{fig:overall}
  \vspace{-.4cm}
\end{figure} 

\textbf{Defining ``Usable XAI''.} In light of the above considerations, in the context of LLMs, we define Usable XAI which includes two aspects.
\textit{(1) Leveraging XAI to improve LLMs and AI Systems.} Beyond just producing explanations of LLMs, we explore whether these explanations can pinpoint issues for model debugging or improve the overall performance of LLMs or AI models, such as accuracy, controllability, and trustworthiness.
\textit{(2) Enhancing the usability of XAI through LLMs.} The human-like communication ability of LLMs can enhance model explanations in terms of user-friendliness, by converting the numerical values into understandable language. Also, the commonsense knowledge stored in LLMs can significantly boost the practicality of existing XAI frameworks, by playing the role of humans and alleviating the need for real human involvement in AI workflows.

\textbf{Contribution of this paper.} 
In this paper, we investigate 10 strategies towards usable XAI techniques in the context of LLMs, as shown in Figure~\ref{fig:overall}.
% As shown in Figure~\ref{fig:overall}, these strategies are organized into two major categories: (1) Usable XAI for LLMs; (2) LLM for Usable XAI. Additionally, 
% We also conducted case studies to support the discussion of selected techniques. 
For each strategy, we also explore the open challenges that require further investigation in future work.
\begin{itemize}[leftmargin=*, topsep=0pt, itemsep=0pt]
\item \textbf{Leveraging XAI to improve LLMs and AI Systems} We introduce how interpretation can be utilized to enhance LLMs and AI models. We follow the LLM development pipeline in reverse order—spanning post-hoc analysis, prompt improvement, and data enhancement. 
First, we investigate how post-hoc explanations could be utilized to diagnose and enhance LLMs. We study three types of post-hoc explanation methods, targeting LLM predictions (\textbf{Section~\ref{sec:attribution}}), LLM components (\textbf{Section~\ref{sec: module_interpret}}), and training samples (\textbf{Section~\ref{sec:sample}}), respectively, followed by how explanations could be leveraged to scrutinize and boost LLM trustworthiness (\textbf{Section~\ref{sec:diagnosis}}), including security, fairness, toxicity, and truthfulness.
Second, we discuss two strategies for designing explainable prompts to guide LLM reasoning: Chain-of-Thought prompts (\textbf{Section~\ref{sec:prompt_cot}}) and knowledge-enhanced prompts (\textbf{Section~\ref{sec:prompt_knowledge}}). Third, we introduce the use of LLM explanations to augment training data toward improving AI models (\textbf{Section~\ref{sec:data}}). 
% we discuss how explainability could guide the augmentation of data, including both inference data (i.e., prompts) and training data. Specifically, 

\item \textbf{Enhancing XAI Usability through LLMs} In this part, we investigate strategies for leveraging the advanced capabilities of LLMs to address the challenges in traditional XAI domains, thus enhancing the usability of XAI in practice. 
First, we examine ways to enhance the user-friendliness of explanations through the generative capabilities of LLMs (\textbf{Section~\ref{sec:user-friendly}}). Second, we introduce how to automate the design of interpretable AI workflows by leveraging the planning abilities of LLMs (\textbf{Section~\ref{sec:architecture}}). 
Third, we introduce how to facilitate the evaluation of XAI methods by utilizing the unique property of LLMs in emulating human cognition processes (\textbf{Section~\ref{sec:mimic_humans}}).
\end{itemize}

\textbf{Differences between this paper and existing surveys.} 
Many surveys have been conducted to examine Explainable AI~\citep{du2019techniques,tjoa2020survey,dovsilovic2018explainable,murdoch2019definitions}. % or Interpretable Machine Learning~\citep{murdoch2019definitions}. 
This paper differs from existing work as we focus on explanation methods for large language models.
Meanwhile, different from the existing survey~\citep{zhao2023explainability,ferrando2024primer} that reviews explanation methods for LLMs, our paper places an emphasis on the XAI usability in LLM studies.
To the best of our knowledge, the most related paper to our survey is~\citep{luo2024understanding}, which also discusses several aspects where explanations can improve LLM performance. Nevertheless, this light-weight investigation lacks a thorough examination of XAI methods (e.g., sample-based explanation, interpretable workflows, explainable prompts) and how LLMs can benefit existing XAI frameworks (e.g., data augmentation, improving user-friendliness, XAI evaluation). 
Finally, our paper contributes further by providing detailed case studies and open-sourced codes, fostering future research in applying explanations effectively within the LLM context.

\section{LLM Diagnosis and Enhancement via Attribution Methods}\label{sec:attribution} 
This section introduces attribution methods as post-hoc explanations for LLMs, and how we can discover model defects with attribution scores. 
We start by reviewing existing attribution techniques and discussing which methods are still suitable for explaining LLMs. 
%Since LLMs widely serve both classification and generation tasks, our discussion categorizes the attribution methods accordingly. 
After that, we explore cases that apply attribution methods to assess LLM-generated output quality. Finally, we discuss future work of designing novel post-hoc explanations for LLMs.

\subsection{Attribution Methods for LLMs}
The attribution-based explanation quantifies the importance of each input feature that contributes to the prediction. 
Given a language model $f$ with a prediction $\hat{y}=f(x)$ according to the $N$-words input prompt $x$, the explainer $g$ assesses the influence of input words in $x$ as: $a = g(x, \hat{y}, f)$, where $a$ stores the importance scores of words. 
%Typically, the sign of $a_n \in \mathbf{a}$ indicates word $x_n$ positively or negatively influences $\hat{y}$, and a greater value of $|a_n|$ indicates a stronger impact. 
In \textit{text classification}, $\hat{y}$ denotes a specific class label. In \textit{text generation}, $\hat{y}$ represents a varying length of generated text.
%Existing attribution-based explanation methods focus on classification tasks and cannot be directly applied to the generation task.
The primary distinction between them is that: classification is limited to a specific set of predictions, while generation encompasses an endless array of possibilities. 
%Typically, a classification model outputs a number between 0 and 1 to indicate its confidence in predicting an instance with a particular label. 
%However, a generative model may express the same prediction with numerous expressions. 
%For instance, in sentiment analysis, a language model can be instructed to output a number between 0 and 1 that indicates the positivity of input text by adding a linear layer and a sigmoid function on top of the language model. 
%However, in the generative setting, the model can express this positivity in numerous expressions, such as ``the reviewer definitely loves this movie'' and ``it is a strong positive movie review''. 
%This distinction poses a unique challenge in adapting explanation methods from classification to generation tasks. 
In the following, we review related works based on the tasks to which they are applicable.

\subsubsection{Attribution Methods for Classification}
\label{sec:attribute_classification} 
Common attribution methods~\citep{du2019techniques,murdoch2019definitions} for deep models include gradient-based methods, perturbation-based methods, surrogate methods, and decomposition methods.
%We introduce the general idea and representative examples for each category, followed by the analysis of their suitability for explaining large language models.
%\textbf{Perturbation-based Explanation.}
\begin{itemize}[leftmargin=*, topsep=0pt, itemsep=0pt]
\item \textbf{Perturbation-based explanation} assesses the importance of input features by perturbing them and monitoring changes in prediction confidence, i.e., $a_n = p(\hat{y}|x) - p(\hat{y}|x_{/n})$, where $x_{/n}$ refers to the input sequence with the $n$-th feature being perturbed or deleted~\citep{li2016visualizing,li2016understanding,wu2020perturbed}. 
%Each feature could refer to a word~\citep{li2016visualizing}, a phrase~\citep{wu2020perturbed}, or a word embedding~\citep{li2016understanding}.
%The underlying principle is that perturbing a more important feature should result in a more pronounced alteration in the model's prediction confidence. 
This approach has limitations, particularly in its assumption that features are independent, which is not always the case with textual data. Additionally, it is computationally intensive for explaining LLMs, requiring $N$ inferences for an input of $N$ words.
%\textbf{Gradient-based Explanation.}
\item \textbf{Gradient-based explanation} offers a computationally efficient approach for estimating model sensitivity to input features based on gradients $\frac{\partial p(\hat{y}|x)}{\partial \mathbf{x}_n}$, where $\mathbf{x}_n$ refers to the embedding of word $x_n$~\citep{li2016visualizing,kindermansreliability,shrikumar2017learning,ebrahimi2018hotflip,mohebbi2021exploring}. 
%Some methods employ the $L_2$-norm of gradients to assess word importance~\citep{li2016visualizing}, i.e., $a_n = \| \frac{\partial p(\hat{y}|x)}{\partial \mathbf{x}_n} \|_2$. 
This approach only requires a single inference and one backpropagation pass, achieving an efficient way to estimate the input word sensitivities. However, it is still expensive to compute a backward pass on such giant LLMs.
%Some extended methods multiply the gradient with the word embedding~\citep{kindermansreliability,ebrahimi2018hotflip,mohebbi2021exploring}, i.e., $a_n = \frac{\partial p(\hat{y}|x)}{\partial \mathbf{x}_n}\cdot \mathbf{x}_n$.
%These methods may yield explanations with limited faithfulness for deep models~\citep{shrikumar2017learning}, as gradients only reflect the local relationship between input variation and output variation. 
%To address this, Integrated Gradients (IG) has been proposed~\citep{sundararajan2017axiomatic,sikdar2021integrated,sanyal2021discretized,enguehard2023sequential}, which accumulates gradients as input transitions from a reference point to the actual data point. Nevertheless, IG entails multiple rounds of inference and backpropagation, thus significantly increasing computational demands.
%\textbf{Surrogate-based Explanation.}
\item \textbf{Surrogate-based explanation} understands the target model $f$ by constructing a simpler model $g$ trained on $\mathcal{D}(x, \hat{y}) = \{(\widetilde{x}_k, \widetilde{y}_k)\}_{k=1}^K$, where %$\mathcal{D}(x, \hat{y})$ denotes a dataset constructed for the target instance $(x, \hat{y})$. Here, 
$\widetilde{x}_k$ is usually obtained by perturbing $x$, and $\widetilde{y}_k=f(\widetilde{x}_k)$~\citep{ribeiro2016should,lundberg2017unified,kokalj2021bert}. The surrogate model $g$, ranging from basic linear models to sophisticated decision trees, serves as a proxy to approximate the decision boundary of $f$ for the target instance $(x,\hat{y})$. 
%Notable examples include LIME~\citep{ribeiro2016should}, SHAP~\citep{lundberg2017unified}, and TransSHAP~\citep{kokalj2021bert}, where the first two are designed for general deep neural networks and the last one is tailored for Transformer-based language models. 
Nevertheless, a significant limitation of this approach is its repeated interactions with the target model, which leads to huge computing requirements. In addition, it is impractical to find a simple and explainable model $g$ that can estimate the advanced LLMs well.
%\textbf{Decomposition-based Explanation.}
\item \textbf{Decomposition-based explanation} assigns linearly additive relevance scores to inputs, effectively breaking down the model's prediction~\citep{montavon2019layer,montavon2017explaining,voita2019analyzing,voita2020analyzing,wu2021explaining}.
%Layer-wise Relevance Propagation~\citep{montavon2019layer} and Taylor-type Decomposition~\citep{montavon2017explaining} are well-known techniques for computing these relevance scores. These methods have been adapted for Transformer-based language models in various research~\citep{voita2019analyzing,voita2020analyzing,wu2021explaining}. 
However, a primary challenge in implementing such an approach is the need for tailored decomposition strategies to accommodate different model architectures. 
%Notablely, even though most LLMs are based on Transformer~\citep{vaswani2017attention}, many key variations still exist. 
%Although many LLMs are based on the Transformer framework, there are key variations between them, such as LLaMA~\citep{touvron2023open} and GPT~\citep{openai2023gpt}, particularly in aspects like positional encoding strategy and feed-forward network design. 
This challenge poses a limitation on the universal applicability of decomposition methods for general-purpose interpretation. 
\end{itemize}

\subsubsection{Attribution Methods for Generation}
\begin{figure*}
    \centering
    \includegraphics[width=0.95\textwidth]{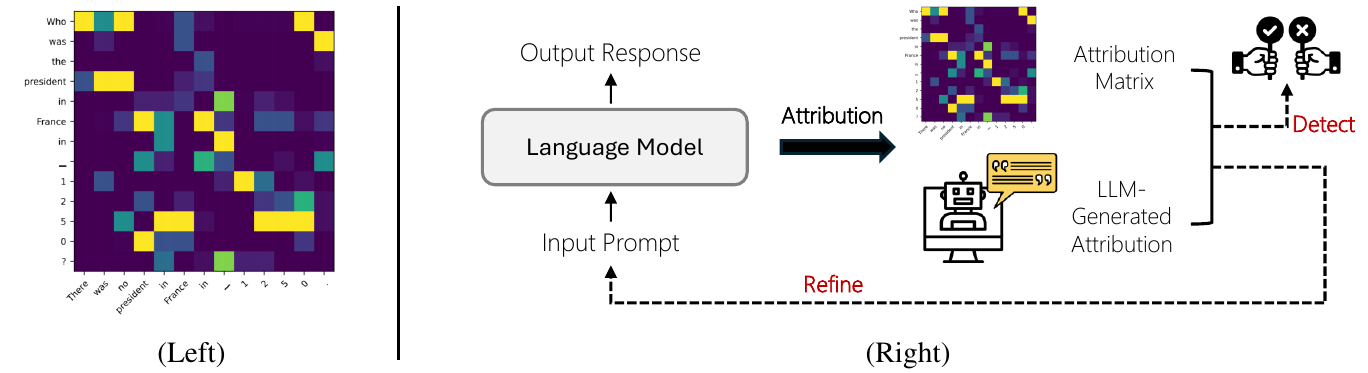}
    \vspace{-0.1cm}
    \caption{Left: An example of attribution saliency maps. Right: The pipeline of making use of attribution results to improve LLMs.}
    \label{fig:attribution_usable}
    \vspace{-0.4cm}
\end{figure*}

With advancements in generative AI, interpreting why an output is generated has become increasingly important.
The explanation of LLM generation can be defined as attributing the overall confidence $p(\hat{y}|x)$ to the input $x$, where $\hat{y}$ denotes the generated response $\hat{y}=[\hat{y}_1,...,\hat{y}_M]$ with $M$ words. 
An approach is to treat the text generation process as a sequence of word-level classification tasks. This perspective allows for the application of \textbf{existing classification-based explanation techniques} to assess the influence of each input word $x_n$ in relation to each output word $\hat{y}_m$.
In particular, \citet{wu2023language} proposes a gradient-based method for input-output attribution. Specifically, the importance score $a_{n,m}$ defined between input token $x_n$ to output token $\hat{y}_m$ can be estimated as:
\begin{equation}
    a_{n,m} = p(\hat{y}_m | x, \hat{y}_{1:(m-1)}) - p(\hat{y}_m | x_{/n}, \hat{y}_{1:(m-1)}) \approx \frac{\partial f(\hat{y}_m | x, \hat{y}_{1:(m-1)})}{\partial \mathbf{E}[x_n]}\cdot\mathbf{E}[x_n] ,
\end{equation}
where $\hat{y}_{1:(m-1)}$ denotes the first $m-1$ tokens of response, $x_{/n}$ means removing the $n$-th input token from $x$, and $\mathbf{E}[x_n]$ is the embedding of token $x_n$.
%, resulting in a corresponding attribution score $a_{n,m}$. 
The total contribution of each input word $x_n$ can be estimated by averaging the attributions $a_{n,m}$ for $m=1,...,M$~\citep{selvaraju2016grad}.
% , denoted as $a_n=\mathrm{Agg}([a_{n,1},...,a_{n,M}])$.
%This is accomplished by aggregating the individual attributions for all output words corresponding to the input word, denoted as $a_n=\mathrm{Aggregate}([a_{n,1},...,a_{n,M}])$. 
% The simplest strategy for aggregation is to average the attributions~\citep{selvaraju2016grad}. 
% However, it is observed that attribution scores $[a_{n,1},...,a_{n,M}]$ of different output words are not comparable. 
%For example, the attribution scores for function words (e.g., ``the'', ``is'', ``have'') are often disproportionately larger than the scores for semantic meaningful words (e.g., verbs and nouns). 
% Therefore, it is necessary to normalize the scores prior to the aggregation.
Figure~\ref{fig:attribution_usable} (left) plots an example, where each index in the Y-axis refers to an input token, while that in the X-axis is an output token. 
A higher attribution score is brighter. 
Given that not all output tokens warrant interpretation, \citet{Qi_2024} outlines a two-step procedure where informative tokens are first identified, and then attribution is made to these tokens using gradient-based or other attribution techniques~\cite{yin2022interpreting}. It also develops a contrastive explanation method, which aims to explain why the model predicted one target token $\hat{y}_m$ over an alternative token $\hat{y}'_m$.
Furthermore, a Python library has been developed to incorporate existing attribution methods for sequential generation of language models~\citep{sarti2023inseq}.

\textbf{Time complexity} is a critical consideration when applying attribution methods to language generation models like LLMs. 
In particular, the perturbation-based and gradient-based approaches may be limited by their huge demand for computing resources, while the surrogate-based and decomposition-based strategies are restricted by their limited generalizability across different model architectures. 
To emphasize the computing costs of applying attribution-based approaches for generative LLMs, we present the time complexity of several representative methods in Table~\ref{table:time_complexity}, by assuming the model takes $N$ words as input and predicts $M$ words as output. 
It shows that existing attribution methods require a large number of forward or backward operations, emphasizing the exploration of efficient methods.

%, so that the scores $[a_{n,1},...,a_{n,M}]$ become comparable for $1\leq m \leq M$. 
%In this example, the user attempts to direct the model to output information that does not exist, namely the French president in 1250. 
% In this example, the model realizes that the French president in 1250 does not exist and refuses to answer. %The model response can be realized as three parts, ``There was no'', ``president in France'', and ``in 1250''. 
% According to the figure, the beginning of the response is generated because of tokens ``Who'' and ``president'', while both ``France'' and ``1250'' are used to generate ``president in France''. Finally, the model emphasizes the date ``1250'' again by referencing corresponding words. 
% This example aligns with human understanding and highlights its usage in the future.   
%However, current research on attribution-based explaining for generative LLMs is still in its early stages, and only a limited number of methods have been proposed.

\begin{table}[t]
\small
\centering
\caption{Time complexity analysis on different attribution methods for the generative task.}
\vspace{-0.4cm}
\label{table:time_complexity}
\begin{tabular}{c|c|c|l}
\toprule
Method & Forward & Backward & Notes \\ \hline
Mask Perturbation & $\mathcal{O}(N)$ & 0 & -  \\ \hline
Gradient$\times$Input & $\mathcal{O}(1)$ & $\mathcal{O}(M)$ & - \\ \hline
Integrated Gradients & $\mathcal{O}(N_{step})$ & $\mathcal{O}(N_{step}\cdot M)$& $N_{step}$ is the number of steps for integrating gradients.  \\ \hline
LIME & $\mathcal{O}(N_{aug})$ & 0 & $N_{aug}$ is the number of augmented samples. \\ \hline
SHAP & $\mathcal{O}(2^N)$ & 0 & - \\ \bottomrule
\end{tabular}
\vspace{-0.6cm}
\end{table}

Another category of attribution approaches leverages the \textbf{generative capabilities of LLMs} to provide citations to one or more text passages that support the output they generate. 
The attribution can be obtained by explicitly prompting LLMs to quote the curated sources of information~\cite{weller2024according}.
The cited contents can originate either from \textbf{external knowledge sources} or from the \textbf{LLMs' own memory}.
For instance, \citet{sunrecitation} introduces RECITE, which first retrieves one or more relevant passages from the LLMs' memory via sampling and then generates the final response based on these passages.
Conversely, many systems assume the presence of an external retrieval corpus. In these cases, the LLM output includes both an answer string and a reference to a short segment of text from the corpus that substantiates the answer~\cite{bohnet2022attributed,gao2023enabling,asaiself}.
Nevertheless, LLMs may hallucinate and produce fabricated information. To address this, various benchmarks have been developed to assess the faithfulness of LLM-generated attributions, i.e., whether the generated statements are fully supported by the cited references. 
For details on evaluation methods, we refer readers to Section~\ref{sec:mimic_humans}.

\subsection{Usability of Attribution Methods for LLMs}
Explanations provide a lens to evaluate whether a model's predictions are grounded in reasonable rationale and to identify opportunities for improving prompt instructions.
Based on this, we discuss two scenarios where LLMs can benefit from attributions: model diagnosis and model enhancement.

\begin{itemize}[leftmargin=*, topsep=0pt, itemsep=0pt]
\item \textbf{Model Diagnosis.} 
By comparing the LLM's output with attribution results, it is possible to detect ungrounded or unsupported information. 
Formally, let $x$ denote the input, $\hat{y}=f(x)$ be the target model's output, and $a$ be the attribution.
The goal is to construct a detector $g_{\mathrm{detect}}(a, x, \hat{y}) \in \{-1, +1\}$ where $+1$ indicates that the prediction can be trusted, while $-1$ means the opposite.
Here $g_{\mathrm{detect}}$ can be implemented as an LLM. 
For instance, \citet{gao2023rarr} proposes a framework called RARR, which leverages attribution to identify unsupported or misleading content generated by LLMs. In particular, the attribution step is automated by using a query generator to produce comprehensive questions about different aspects of the text that need verification.
In addition, \citet{agrawal2024language} proposes "consistency checks", which asks the LLM to perform attribution multiple times for the same output. Inconsistencies in the attribution results serve as indicators of potential issues with the factuality of the LLM's output.

\item \textbf{Model Enhancement.} 
Attribution results can serve as additional information to refine prompts, thereby improving LLM performance.
For instance, \citet{sunrecitation} proposes "recitation" as a new two-step approach where an LLM $f$ first explicitly recalls relevant factual knowledge $a$ from its own memory before generating an answer for input $x$. The LLM inference is augmented as $\hat{y} = f(x, a)$ instead of the standard $f(x)$. This recitation step allows the model to better access and verify its stored knowledge, similar to how a student might first recite relevant facts they've learned before answering an exam question, leading to more accurate responses.
In addition, \citet{krishna2024post} proposes a framework called AMPLIFY, which uses attribution results to improve LLM performance by crafting effective prompts. It selects samples in the validation set that were misclassified by the LLM, and uses another model (e.g., GPT-2 or BERT) to compute their attribution scores with respect to the ground truth labels. These samples and their corresponding explanations are used to construct prompts for future test samples.
\end{itemize}

\subsection{Challenges}
% \subsubsection{Novel Explanation Paradigms are Needed for LLMs} 
We discuss two major challenges regarding attribution methods for explaining LLMs, including \textbf{efficiency} and \textbf{faithfulness}.
First, according to our analysis in Table~\ref{table:time_complexity}, many traditional attribution methods are computationally expensive when adapted to LLMs. Furthermore, these methods were primarily developed for tasks in the context of \textit{classification or regression}, while the more urgent need lies in designing attribution methods tailored for LLM \textit{generation}. The diverse and dynamic nature of LLM generation demands the development of novel explanation paradigms that go beyond existing frameworks.
Second, while some approaches utilize the generative capabilities of LLMs to produce free-text attributions, the faithfulness of these attributions remains underexplored. Ensuring that the generated explanations truly reflect the model's reasoning process is a challenge to be tackled for building trust and reliability in LLM explanations.

\tikzstyle{my-box}=[
rectangle,
draw=black,
rounded corners,
text opacity=1,
minimum height=1.5em,
minimum width=5em,
inner sep=2pt,
align=center,
fill opacity=.5,
]
\tikzstyle{leaf2}=[
my-box, 
minimum height=1.5em,
fill=blue!10, 
text=black,
align=left,
font=\tiny,
inner xsep=2pt,
inner ysep=3pt,
text width=14.5em,
]

\begin{figure*}[t]
\centering
\resizebox{0.85\textwidth}{!}{
\begin{forest}
forked edges,
for tree={
    grow=east,
    reversed=true,
    anchor=base west,
    parent anchor=east,
    child anchor=west,
    rectangle,
    draw,
    font=\tiny,
    rounded corners,
    align=center,
    minimum width=4em,
    edge+={darkgray, line width=1pt},
    s sep=3pt,
    inner xsep=2pt,
    inner ysep=2pt,
    ver/.style={rotate=90, child anchor=north, parent anchor=south, anchor=center, text centered},
},
[Interpreting Model Internals,ver,fill=gray!20,text centered,text width=10em
  [Explanation of\\Model Components,ver,fill=orange!40,text centered,text width=7em
    [Importance Analysis on \\ Single Component,fill=orange!10,text centered,text width=6em
      [{Contextual Impacts: \cite{geva2022transformer}, \cite{ferrando2023explaining}, \cite{wanginterpretability}, \cite{elhage2021mathematical}, \cite{meng2022locating}, \cite{wang2022interpretability}, \cite{mohebbi2023quantifying}.}, leaf2, align=left]
      [{Global Impacts: \cite{geva2021transformer}, \cite{dar2023analyzing}, \cite{wu2023language}.}, leaf2, align=left]
    ]
    [Circuit Analysis on \\ Multiple Components,fill=orange!10,text centered,text width=6em
      [{\raggedleft\cite{cammarata2020thread}, \cite{wang2022interpretability}, \cite{hanna2023does}, \cite{geiger2021causal}, \cite{conmy2023towards} \cite{sarti2023inseq}, \cite{hanna2024have}, \cite{marks2024sparse}.}, leaf2, align=left]
    ]
    [Usability,fill=orange!10,text centered,text width=6em
     [{Knowledge Editing: \cite{dai2022knowledge}, \cite{wang2022finding}, \cite{elhage2021mathematical}, \cite{meng2022locating}, \cite{meng2022mass}, \cite{hase2024does}, \cite{li2024pmetprecisemodelediting}, \\ \cite{gupta2024unifiedframeworkmodelediting}, \cite{wu2023depndetectingeditingprivacy}, \cite{wang2024detoxifyinglargelanguagemodels}.}, leaf2, align=left]
     [{Model Pruning: \cite{dalvi2020analyzing}, \cite{men2024shortgptlayerslargelanguage}, \cite{gromov2025unreasonableineffectivenessdeeperlayers}, \cite{dalvi2020analyzing}, \cite{fan2024layersllmsnecessaryinference}.}, leaf2, align=left]
    ]
  ]
  [Explanation of\\Latent Representations,ver,fill=orange!40,text centered,text width=7em
    [Non-Training\\based Methods,fill=orange!10,text centered,text width=6em
      [{Projecting $\mathbf{x}^l$ to Output Embeddings: \cite{2020logitlens}, \cite{jastrzebski2018residual}, \cite{belrose2023eliciting}, \cite{din2024jump}.}, leaf2, align=left]
      [{Feeding $\mathbf{x}^l$ as Input Embeddings: \cite{chenselfie}, \cite{dumas2024separating}, \cite{pan2024latentqa}.}, leaf2, align=left]
    ]
    [Supervised Training\\based Methods,fill=orange!10,text centered,text width=6em
      [{Identifying Concept Existence: \cite{hewitt2019structural},\cite{chenprobing}, \cite{liu2019linguistic}, \cite{hewitt2019structural}, \cite{burns2022discovering},\\ \cite{zou2023representation}, \cite{macdiarmid2024simple},\cite{zhengprompt}, \cite{hewitt2019designing}, \cite{pimentel2020information}, \cite{voita2020information}.}, leaf2, align=left]
      [{Measuring Concept Importance: \cite{kim2018interpretability}, \cite{wei2021analysing}, \cite{bai2022concept}, \cite{nicolson2024textcavs}.}, leaf2, align=left]
    ]
    [Unsupervised Training\\based Methods,fill=orange!10,text centered,text width=6em
      [{Orthogonal Assumption on Concepts: {\cite{millidge2022singular}, \cite{wu2023language}}.}, leaf2, align=left]
      [{Sparse Assumption on Concepts: \cite{olshausen1997sparse}, \cite{cunningham2023sparse}, \cite{brickentowards}.}, leaf2, align=left]
    ]
    [Usability,fill=orange!10,text centered,text width=6em
      [{Data Pre-processing: \cite{ge2024clusteringrankingdiversitypreservedinstruction}, \cite{xu2023wizardlmempoweringlargelanguage}, \cite{li2024quantityqualityboostingllm}, \cite{yang2025diversitydrivendataselectionlanguage}.}, leaf2, align=left]
      [{Training-Time Regularization: \cite{yin2024directpreferenceoptimizationusing}, \cite{wu2025selfregularizationlatentspaceexplanations}.}, leaf2, align=left]
      [{Inference-Time Control: \cite{wu2025interpreting}, \cite{huang2024steering}, \cite{stolfo2024improving}, \cite{rimsky2024steering}, \cite{arditi2025refusal}, \cite{bhattacharjee2024towards},\\ \cite{zou2023representation}, \cite{rimsky2024steering}, \cite{chen2023inside}, \cite{ahdritz2024distinguishing}, \cite{singh2024mimic}, \cite{zou2023representation}, \cite{scialanga2025sake}.}, leaf2, align=left]
    ]
  ]
]
\end{forest}
}
\vspace{-0.2cm}
\caption{Taxonomy of explanation methods for LLM internals, including explaining \textit{model components} and \textit{hidden representations}.}
\vspace{-0.4cm}
\label{fig:taxonomy_sec3}
\end{figure*}

% \begin{figure*}
%     \centering
%     \includegraphics[width=0.89\textwidth]{Figures/ModelExplain_v4.pdf}
%     \vspace{-0.2cm}
%     \caption{Review of interpretation methods for LLM components and their applications. We categorize methods according to the target LLM modules: self-attention layers and feed-forward layers.}
%     \label{fig:enter-label}
%     \vspace{-0.2cm}
% \end{figure*}

\section{LLM Diagnosis and Enhancement via Interpreting Model Internals} \label{sec: module_interpret}
This section discusses XAI methods that interpret LLM internals. We delve into the insights these methodologies offer, which can be instrumental in refining and enhancing the design of language models. 
Specifically, we review research that focuses on interpreting \textbf{model components} and \textbf{latent representations}, respectively, as shown in~Figure~\ref{fig:taxonomy_sec3}.

\subsection{Explanation of Model Components} 
LLMs adopt transformers as their basic architectures, which typically comprise two types of major components: Self-attention Mechanism (ATT) and Feed-forward Network (FFN). 
Researchers have focused on analyzing individual components as well as their interactions to gain insights into how LLMs make their predictions. 
We start by reviewing the connection between these components, focusing on how information is transformed through the residual stream. 
We then discuss methods to analyze the contribution of single or multiple components, respectively. 
We conclude by summarizing how these methods can be applied to improve model performance.

\subsubsection{\textbf{Linear Relation between Components}} 
\label{sec:linear_structure}
We consider a decoder-only LLM $f$ with $L$ internal layers, each of which is either an ATT or FNN layer, denoted as $f^l$. 
There is an input embedding matrix $\mathbf{W}_\text{e}$ that encodes each word of input text $x$ into a $D$-dimensional vector, i.e., $\mathbf{x}^0=\mathbf{W}_\text{e}[x]\in\mathbb{R}^{|x|\times D}$. 
The internal representation at each layer is computed by $\mathbf{x}^l=f^l(\mathbf{x}^{l-1}) + \mathbf{x}^{l-1}$. 
Finally, there is an output embedding matrix $\mathbf{W}_\text{u}$ that decodes the hidden representations for next word prediction, i.e., $f(x)=\mathbf{x}^L\cdot\mathbf{W}_\text{u}$. 
By decomposing the final prediction~\cite{elhage2021mathematical}, we have 
\begin{equation}
f(x)
~=~
\underbrace{ \mathbf{x}^0\,\mathbf{W}_\text{u} }_{\text{Residual Stream}}
~+~\sum_{l=1}^L 
\underbrace{ f^l\Bigl(\mathbf{x}^{l-1}\Bigr)\,\mathbf{W}_\text{u}}_{\text{Layer Update}}. 
\label{eq:model_component}
\end{equation}
Equation~\eqref{eq:model_component} demonstrates two critical insights in understanding LLM mechanism: 
\textbf{(1) Model prediction is a sum of layer outputs~\cite{mickus2022dissect}.} That is to say, the final prediction is an additive combination of the initial input embedding and the contribution from each layer. This linear transformation allows us to directly trace how each component shapes the overall prediction. 
Please note that the residual mechanism does not necessarily lead to a linear structure of model hidden spaces. For example, ResNet~\cite{he2016deep} does not show such a linear structure because there are non-linear operations applied to its residual stream. 
\textbf{(2) The residual stream serves as a communication channel across layers~\cite{elhage2021mathematical}.} In particular, each layer first extracts information from the residual stream, and then injects its processed output back into it. This shared residual stream enables multiple layers to work collaboratively, indirectly contributing to specific tasks by continuously updating the shared residual stream.

\subsubsection{\textbf{Importance Analysis on Single Component}} 
\label{single_component}
Inspired by the first insight of Equation~\eqref{eq:model_component}, early works~\cite{geva2022transformer,ferrando2023explaining,wanginterpretability} quantify the \textbf{contextual impact} of a single layer $f^l$ in predicting a certain word $w$ under the context $x$ by computing $f^l(\mathbf{x}^{l-1})\,\mathbf{W}_\text{u}[:, w]$, where a greater value indicates a stronger impact.  
Following this trend, some researchers~\cite{elhage2021mathematical,meng2022locating} propose to more precisely quantify the contextual impact by conducting interventions on layer outputs, i.e., $\text{diff}(\mathbf{h}\,\mathbf{W}_\text{u}[:, w],\widetilde{\mathbf{h}}\,\mathbf{W}_\text{u}[:, w])$, where $\text{diff}(\cdot,\cdot)$ measures the difference of two inputs, $\mathbf{h}=f^l(\mathbf{x}^{l-1})$, and $\widetilde{\mathbf{h}}$ is a intervention of $\mathbf{h}$. 
In practice, the choices of $\widetilde{\mathbf{h}}$ can vary, such as a constant vector~\cite{wang2022interpretability,mohebbi2023quantifying}, a noise vector~\cite{meng2022locating}, and a hidden representation of other context~\cite{hanna2023does}. 
On the other hand, researchers~\cite{geva2021transformer,dar2023analyzing,wu2023language} extend this framework to analyze the \textbf{global impact} of a single layer $f^l$ in predicting word $w$ by computing $\mathbf{W}^l_\text{o}\cdot\mathbf{W}_u[:, w]$, where $\mathbf{W}^l_\text{o}$ is the output weight matrix of layer $f^l$ that maps the processed results back to the residual stream.

\subsubsection{\textbf{Circuit Analysis on Multiple Components}} 
\label{multiple_components}
Motivated by many emerging abilities of LLMs, it is interesting to explore how \textbf{different layers work collaboratively} to perform such complex tasks, called circuit analysis~\cite{cammarata2020thread}. 
Specifically, the goal of circuit analysis is to identify a subset of layers, whose combination performs a particular function, such as ``indirect objects identification''~\cite{wang2022interpretability} and ``number comparison''~\cite{hanna2023does}. 
Recall that the residual stream enables the communication across layers, researchers~\cite{geiger2021causal} formalize the forward pass of an LLM as a directed acyclic graph, where each node is one particular layer, and an edge is a representation flow from one layer to another via the residual stream. 
Under this setting, a circuit is a sub-graph with a distinct function. 
Identifying a specific circuit consists of three steps~\cite{conmy2023towards}: (1) selecting an interested task and constructing a dataset, (2) building the model’s graph, and (3) measuring the importance of sub-graphs with interventions (Sec.~\ref{single_component}). 
This circuit identification process is time-consuming as it needs a large number of forward passes for interventions. 
To overcome this challenge, researchers~\cite{sarti2023inseq,hanna2024have,marks2024sparse} propose a gradient-based method to approximate the intervention results with one more backward pass for each sample.

\subsubsection{\textbf{Usability of Component Analysis}} 
\label{usability_sec31}
The success of component analysis enables researchers to isolate the entire LLM into different parts for specific tasks. 
Therefore, the direct application of this technique is either modifying (enhancing or weakening) or removing certain parts that are essential or redundant to downstream tasks, resulting in two common applications: Knowledge Editing and Model Pruning.

\begin{itemize}[leftmargin=*, topsep=0pt, itemsep=0pt]
\item \textbf{Knowledge Editing.} 
Component analysis can be used to locate the exact position storing a certain piece of knowledge~\cite{dai2022knowledge,wang2022finding}, and then update it accordingly to control the minimum effect on other knowledge.  
Specifically, we aim to update an outdated knowledge $t=(s,r,o)$ to a new one $t'=(s,r,o')$. 
Given the prefix $[s,r]$ of $t$, ROME~\cite{meng2022locating} first identifies a specific layer $f^{l*}\in f$ whose activation maximally contributes to predicting the original object $o$. 
This is done by maximizing the difference $f(o|[s,r]) - f(o|[s,r], do(\mathbf{h}^l = \mathbf{h}^l + \epsilon))$, where $\mathbf{h}^l$ denotes the output activation for input $[s,r]$ at layer $f^l$, and $\epsilon$ denotes a random noise to simulate a perturbation. 
ROME then modifies only the parameters of this identified layer $f^{l*}$ to update the prediction to the new object $o'$, without affecting unrelated knowledge stored elsewhere.
The key limitation of ROME is that it can update only one piece of knowledge at a time, leading to an unacceptable time complexity to update a batch of knowledge. 
To overcome this challenge, follow-up works~\cite{meng2022mass,hase2024does,li2024pmetprecisemodelediting,gupta2024unifiedframeworkmodelediting} scale up this process by skipping the knowledge locating step in ROME, and instead, they focus on learning the representation of updated knowledge at each layer. 
This line of work has been further applied to edit user personal information for privacy~\cite{wu2023depndetectingeditingprivacy}, and edit out harmful concepts for safety~\cite{wang2024detoxifyinglargelanguagemodels}.

\item \textbf{Model Pruning.}
Component analysis allows researchers to identify the specific roles of each model component for a given task, and thus, redundant or irrelevant components can be pruned to speed up the inference. 
\citet{dalvi2020analyzing} directly measure the functionality of $f^l\in f$ by monitoring the expected difference $\mathbb{E}_{x,y\in\mathcal{D}}[p(y|x;f)-p(y|x,do(f^l(l)=\mathbf{0});f)]$, where $\mathcal{D}$ is a collected dataset for the downstream task. 
Recent studies~\cite{men2024shortgptlayerslargelanguage,gromov2025unreasonableineffectivenessdeeperlayers} directly measure the redundancy by quantifying the similarities of hidden representations from different layers. 
Empirical studies~\cite {dalvi2020analyzing,fan2024layersllmsnecessaryinference} found that even over half of the model parameters are redundant for a certain task, and dropping them off can significantly reduce the inference time without sacrificing performance. 

\end{itemize}

\subsection{Explanation of Latent Representations}
Many researchers have studied latent representations of LLMs (i.e., $\mathbf{x}^l$) to interpret model behaviors. 
It is critical to recognize that the representations are not naturally interpretable because of their \textbf{\textit{polysemantic}} nature~\citep{arora2018linear,scherlis2022polysemanticity,brickentowards}, indicating that each dimension of the latent space represents multiple semantic meanings, called \textbf{concepts}. 
Thus, we \textbf{cannot} clearly understand a latent representation by reviewing the values in each dimension. 
There are various lines of work to tackle this challenge, and we categorize them according to whether they require additional training. 
% Some of these methods may heavily rely on the linear structure assumption of LLMs (see Section~\ref{sec:linear_structure}). 

\subsubsection{\textbf{Non-Training based Methods}}
\,

\noindent 
Both the \textit{input and output layer} of transformer-based LLM $f$ are directly interpretable as they operate on human-readable text. Hidden representations across different layers share the same residual stream (see Section~\ref{sec:linear_structure}), allowing information to flow seamlessly between layers. Consequently, non-training based interpretation methods leverage this insight by transferring the hidden representation $\mathbf{x}^l$ to either the first or the last layer to facilitate interpretation.
\begin{itemize}[leftmargin=*, topsep=0pt, itemsep=0pt]
 
\item \textbf{Projecting $\mathbf{x}^l$ to Output Embeddings.} 
Given a hidden representation $\mathbf{x}^l$ at layer $l$, \emph{Logit Lens}~\citep{2020logitlens} understands its semantics by projecting it to the output word embedding matrix 
 $\mathbf{W}_\text{u}$. 
Specifically, we interpret $\mathbf{x}^l$ by collecting the $K$ words whose output embeddings could maximally activate $\mathbf{x}^l$, i.e., 
\begin{equation}
\mathcal{I}=\underset{\mathcal{V}^\prime\subset\mathcal{V},|\mathcal{V}^\prime|=K}{\arg\max}\sum_{w\in\mathcal{V}^\prime}\mathbf{x}^l\cdot \mathbf{W}_\text{u}[:, w],    
\end{equation}
where $\mathcal{V}$ is a pre-defined vocabulary set. 
The Logit Lens can be viewed as an instance of skipping all later layers after $\mathbf{x}^l$~\cite{jastrzebski2018residual}. 
However, researchers~\cite{belrose2023eliciting,din2024jump} find that there is a distribution shift between the intermediate layer $l$ and the last layer $L$, and thus, they propose to train a \textit{translator} to learn the distribution shift to improve its interpretability for representations from shallow layers.

\item
\textbf{Feeding $\mathbf{x}^l$ as Input Embeddings.} 
In contrast with moving hidden representation $\mathbf{x}^l$ to the output layers, \emph{Self-IE}~\cite{chenselfie} proposes feeding $\mathbf{x}^l$ as an input word embedding in another round of model inference, where we prompt LLM to explain the meaning of the fed representation. 
For example, we may instruct LLM to interpret $\mathbf{x}^l$ with the prompt ``Please summarize [X] in one sentence.'', where ``[X]'' is a placeholder, whose embedding will be replaced with our interested hidden representation $\mathbf{x}^l$. 
Since LLMs are trained to follow human instructions, they are expected to generate a sentence to describe the information encoded in $\mathbf{x}^l$. 
\end{itemize}

\subsubsection{\textbf{Supervised Training based Methods}} 
\label{sec:supervised}
\,

\noindent The supervised training-based interpretation methods break the polysemantic nature of LLM latent space by leveraging some \textit{annotated} datasets. 
They focus on exploring two main research questions: (1) whether the LLM latent space encodes a specific concept; and if so, (2) whether this concept impacts LLM's predictions and how significant. 

\begin{itemize}[leftmargin=*, topsep=0pt, itemsep=0pt]
\item \textbf{Identifying Existence of Certain Concepts.} \textit{Probing}~\cite{gupta2015distributional} is the most traditional technique in analyzing the knowledge encoded within hidden representations of LLMs. 
In specific, the Probing technique requires an annotated dataset to train a probe $p$ with parameter $\theta$ to predict the occurrence of interested knowledge, i.e., $p: \mathbb{X}\rightarrow[0,1]$, where $\mathbb{X}$ is the latent space of LLMs at a certain layer.
Here, the dataset consists of positive samples (with label ``1'') that clearly demonstrate our interest concept and negative samples that clearly exclude such concept. 
For example, early works~\cite{belinkov2017neural} use this technique to identify some morphology concepts (e.g., part-of-speech tags and morph tags) in the LLM's latent space by constructing a dataset where each example has human annotations on these tags.  
If the probe $p$ achieves a significantly high accuracy on a certain dataset, then we could conclude that the latent representation encodes that concept. 
This technique has been used to study diverse concepts, ranging from low-level linguistic patterns~\cite{belinkov2017neural,hewitt2019structural,chenprobing,liu2019linguistic,hewitt2019structural} to high-level semantic meanings~\cite{burns2022discovering,zou2023representation,macdiarmid2024simple,zhengprompt}. 
The improvements in the probing technique~\cite{hewitt2019designing,pimentel2020information,voita2020information} mainly focus on designing baselines to confirm that the high accuracy achieved by probe $p$ is led by the knowledge provided by the latent space $\mathbf{x}$. 

\item \textbf{Measuring Importance of Certain Concepts.} 
To quantify how much a specific concept impacts the model’s predictions, researchers often use \emph{TCAV}~\citep{kim2018interpretability}.
We first train a linear probe $p$ with parameter $\theta \in \mathbb{R}^D$ to distinguish latent representations containing concept $c$ or not, where $\theta$ is named as the concept activation vector (CAV). 
TCAV measures how sensitive the model’s output is to perturbations along this concept direction (e.g., via directional derivatives of the model’s logits).
Intuitively, if the model’s predictions change substantially when the representation is shifted along the learned concept direction, we can conclude that the concept has a high \emph{importance} for the model’s decision. 
Mathematically, the perturbation is realized by computing the gradient of $f(x)$ with respect to $\mathbf{x}^l$ and projecting it onto $\theta$ as $\Delta = \nabla_{\mathbf{x}^l} f(x) \cdot \theta$, so that a larger $|\Delta|$ indicates that small shifts in the latent space along $\theta$ lead to substantial changes in $f(x)$, underscoring the concept's influence.

\end{itemize}

\subsubsection{\textbf{Unsupervised Training based Methods}}
\label{unsupervised_hidden_representation}
\,

\noindent
Breaking the polysemantic constraint for interpretation can be formalized as an unsupervised learning problem on the basis vectors of the LLM latent space~\cite{fel2023holistic}, where each basis vector refers to a clear and concise \textbf{semantic concept}, indicating a \textit{monosemantic} nature.  
Mathematically, given a total of $N$ hidden representations $\mathbf{X}\in\mathbb{R}^{N\times D}$, we aim to learn two matrices $\mathbf{A}\in\mathbb{R}^{N\times C}$ and $\mathbf{C}\in\mathbb{R}^{C\times D}$ by minimizing the loss 
\begin{equation}
\underset{\mathbf{A},\mathbf{C}}{\arg\min}\,\,\|\mathbf{X}-\mathbf{A}\mathbf{C}\|^2,   
\end{equation}
where $\mathbf{C}$ serve as the learned concepts, and $\mathbf{A}$ is the coefficient of the instances on each concept. 
There are some additional conditions to ensure that the learned concept vectors satisfy our expectations on monosemantic. 

\begin{itemize}[leftmargin=*, topsep=0pt, itemsep=0pt]
\item
\textbf{Orthogonal Assumption on Concepts.} 
Early works~\cite{millidge2022singular,wu2023language} applies the \textit{Singular Vector Decomposition} to learn a set of concept vectors $\mathbf{C}$ that are perfectly orthogonal, i.e., 
\begin{equation}
    \mathbf{C}\mathbf{C}^\top=\mathbf{I},
\end{equation} 
where $\mathbf{I}$ is an identical matrix. 
However, satisfying this hard request is computing costly in practice, leading to limited scalability in interpreting hidden representations from giant LLMs with both large $N$ and $D$. 

\item
\textbf{Sparse Assumption on Concepts.}
\textit{Sparse Autoencoder}~\cite{olshausen1997sparse} (SAE) is another alternative that relaxes the orthogonal condition to be a smoother one in which the sample coefficients matrix is asked to be sparse, i.e., 
\begin{equation}
    \mathbf{A}=\text{Top-K}(\mathbf{X}\cdot\mathbf{C}^\top), 
\end{equation}
where $\text{Top-K}$ activation enforces all values to be zero unless the $K$ largest ones from the inputs.
Specifically, researchers~\cite{cunningham2023sparse,brickentowards} train a SAE $g(\mathbf{x})=\text{Top-K}(\mathbf{x}\cdot\mathbf{W}^\top)\cdot\mathbf{W}$ to reconstruct hidden representations $\mathbf{x}$ by minimizing $\|\mathbf{x}-g(\mathbf{x})\|^2$, where $\mathbf{W}\in\mathbf{R}^{C\times D}$ and $\text{Top-K}$ activation enforces all values to be zero unless the $K$ largest ones from the input. 
This approach has shown effectiveness in decomposing the latent representations from giant LLMs with a hundred billion parameters, such as GPT-4~\cite{gao2024scaling} and Claude 3~\cite{templeton2024scaling}.  
\end{itemize}

\noindent
%\textbf{Interpreting Extracted Concept Vectors.}
Once these concept vectors are obtained, their representing meanings can be further described with natural language by selecting $M$ input texts that could maximally activate the related concept vectors~\cite{brickentowards}, i.e., 
\begin{equation}
\mathcal{I}_c=\arg\max_{\mathcal{X}^\prime\subset\mathcal{X},|\mathcal{X}^\prime|=M}\sum_{x\in\mathcal{X}^\prime} f^l(x)\mathbf{C}[c].
\end{equation}
However, the latest research~\cite{wu2025interpreting,gur2025enhancing} has shown that such an input-based explanation cannot well interpret their impacts on controlling LLM responses, and thus, they propose to interpret these learned concept vectors by building the connection to the output texts. We refer audiences to check~\cite{shu2025surveysparseautoencodersinterpreting} for a more detailed review of designing and interpreting SAEs for LLMs.

\subsubsection{\textbf{Usability of Interpreting Latent Representations}} 
\label{usability_sec32}
\,

\noindent Latent space interpretations improve LLMs further at each stage of their entire life-cycles. 
In particular, we consider three stages of model development, including the ``pre-processing'' stage for data preparation, the ``training'' stage that trains LLM on prepared data, and the ``inference'' stage that uses a trained model to operate user requests.   

\begin{itemize}[leftmargin=*, topsep=0pt, itemsep=0pt]
\item \textbf{Data Pre-processing.} 
Existing studies have empirically found that the same model but trained on different datasets with different diversity~\cite{ge2024clusteringrankingdiversitypreservedinstruction}, complexity~\cite{xu2023wizardlmempoweringlargelanguage}, and quality~\cite{li2024quantityqualityboostingllm} levels may lead different downstream task performance.  
The latent space interpretation techniques enable researchers to understand the geometric structure of a dataset and, further, to select a subset that preserves a certain geometric structure. 
For example, to obtain a subset of the dataset preserving as diverse topics as possible, SAE-GreedSelect~\cite{yang2025diversitydrivendataselectionlanguage} greedily select examples to maximize the total number of activated SAE features. 
Results show that a selected subset with only 5\% of total samples can train a model with comparable performance to a full-trained one, significantly reducing the training cost.  

\item \textbf{Training-Time Regularization.}
Either the trained parameter $\theta$ of probe $p$ or each column vector $\mathbf{C}[c]$ learned by SAE or SVD indicates a feature in the LLM latent space. 
Therefore, these learned feature vectors can be used to perform feature engineering in the LLM latent space for model training. 
For example, \citet{yin2024directpreferenceoptimizationusing} propose a feature-level constrained fine-tuning approach. This method effectively prevents the fine-tuned model from changing significantly from its pretrained counterpart without explicitly involving the pretrained model during fine-tuning, thereby substantially reducing computational costs. 
In contrast, \citet{wu2025selfregularizationlatentspaceexplanations} introduce a more semantically meaningful regularization approach for model training. They first determine whether each learned feature vector $\mathbf{C}[c]$ is causally relevant to the downstream task by analyzing its natural language explanation $\mathcal{I}_c$. Subsequently, they impose a penalty during fine-tuning to discourage reliance on features identified as task-irrelevant (i.e., shortcuts). This approach effectively enhances the generalizability of the fine-tuned model.

\item \textbf{Inference-Time Control.} 
As the learned feature vectors by probing or SAEs break the polysemantic nature and achieve monosemantic, it is possible to directly steer LLM latent space toward a specific direction without sacrificing other abilities.  
Formally, let $\mathbf{z}$ denote the learned feature vector to an interested concept, we can \textit{erase} it from the hidden representation of input $x$ by $\mathbf{x}^\prime=\mathbf{x}-\text{ReLU}(\mathbf{x}\cdot\mathbf{z}^\top)\cdot\mathbf{z}$, or constantly \textit{amplify} it to a certain level $\alpha$ by $\mathbf{x}^\prime=\mathbf{x}+\alpha\cdot\mathbf{z}$. 
The combination of these two techniques on different categories of feature vectors leverages model steering for different purposes. 
For example, \citet{wu2025interpreting} observe that certain feature vectors semantically correlate with specific safety-related concepts, such as ``First Aid,'' based on summarizing their natural language explanations $\mathcal{I}_c$. 
By enforcing a constant activation of these safety-related feature vectors during model generation, they demonstrate that the model reliably switches from generating harmful responses to explicitly rejecting inappropriate or malicious user inputs. 
Generally, this technique has shown success in steering LLMs during inference to improve their helpfulness~\cite{huang2024steering,stolfo2024improving,rimsky2024steering, he2025saif}, safety~\cite{arditi2025refusal,bhattacharjee2024towards,zou2023representation,rimsky2024steering}, faithfulness~\cite{chen2023inside,ahdritz2024distinguishing}, fairness~\cite{singh2024mimic,zou2023representation}, and currentness~\cite{scialanga2025sake}.

\end{itemize}

\subsection{Challenges} 
%Interpreting the functionality of internal modules is still in its infancy, and we identify two challenges in this direction. 

%\subsubsection{Complexity of Individual Models and Their Interactions}
% circuit -> expert, capsule 
\subsubsection{Explanation Efficiency} 
The most significant challenge in analyzing the inner workings of LLMs is their demand for a large amount of computing resources. 
Circuit analysis (Section~\ref{multiple_components}) is one of the examples that is bounded by this bottleneck. 
Specifically, performing circuit analysis on one sample still requires a large amount of forward inference of LLMs under different perturbed internals of that sample, leading to limited scalability to analyze over a large dataset.
Therefore, existing works on \textbf{circuit analysis focus on studying particular behaviors of LLMs}, such as ``Name Mover Head'' and ``Duplicate Token Head'' for object identifications~\citep{wang2022interpretability}, ``Single Letter Head'' and ``Correct Letter Head'' for multiple-choice question answering~\citep{lieberum2023does}, and ``Capitalize Head'' as well as ``Antonym Head'' for writing~\citep{todd2023function}. 
On the other hand, although SAEs (Section~\ref{unsupervised_hidden_representation}) have shown strong promise in understanding LLM latent representations, training such proxy models requires collecting LLM representations over a large corpus. For example, researchers~\cite{gao2024scaling} found that \textbf{training an SAE with only 1 million features requires over 10 billion training tokens}.
Future works may explore computing-efficient methods to understand LLMs' diverse behaviors comprehensively.
\subsubsection{Explanation Faithfulness}
%Interpreting the functionality of a single neuron (one row/column vector of a weight matrix) fails in analyzing large language models since a single neuron could be activated by multiple and diverse meanings, called polysemantic~\citep{arora2018linear,scherlis2022polysemanticity,brickentowards}.
%This nature leads to poor interpretability: explaining a single neuron usually does not reflect a concise human concept. 
% Some researchers~\citep{elhage2022toy,sharkey2022taking} assume that this phenomenon is caused by the superposition of an over-complete set of features learned by the models.  
% Based on this assumption, we may reach another level of explanation by decomposing the model weights to reconstruct a large number of features. 
Although many proposed methods have shown strong promise with certain usability cases (Section~\ref{usability_sec31} and Section~\ref{usability_sec32}), some researchers still hold concerns about whether they faithfully interpret the inner workings of LLMs since they may not practically outperform some trivial baselines. 
For example, although many researchers~\cite{wu2025interpreting,brickentowards} can apply SAEs (Section~\ref{unsupervised_hidden_representation}) to steer LLM responses for specific purposes, some research~\cite{wu2025axbench,durmus2024steering,bricken2024using} has observed that \textbf{SAE-based methods cannot significantly outperform prompting baselines}, posing a serious concern about the effectiveness of SAEs.
Furthermore, other researchers~\cite{heap2025sparse,paulo2025sparse} suspect that SAEs cannot faithfully explain LLMs' internals, as they observe that \textbf{SAEs can even learn features from randomized LLMs}. 
These concerns suggest two critical questions for future works: (1) How do we ensure our learned features faithfully represent LLM internals? 
(2) In which cases are SAE-based methods more effective than prompting strategies?
%(2) How do we interpret our reconstructed features with human language? 

\section{LLM Debugging with Sample-based Explanation} 
\label{sec:sample}
In this section, we discuss sample-based explanation strategies for LLMs, which aim to trace back the answers generated by LLMs to specific training samples (i.e., documents) or document fragments in the corpora. The utility of sample-based explanations for LLMs is multifaceted. First, tracing back the predictions of LLM to the training samples can provide evidence for the generation results, which facilitates model debugging in cases of errors and increases the trustworthiness of the model from users when the outcomes are accurate. 
Second, it can also help researchers understand how LLMs generalize from training samples. If the outputs of LLMs can be traced back to exact subsequences directly spliced from the training data, it might suggest that the LLM is simply memorizing the data. %In contrast, if the generation results and the influencing training samples are abstractly related, it could indicate that LLMs can understand the concepts and generate responses by reasoning from input prompts. 

% In this section, we start by systematically reviewing traditional sample-based explanation strategies, including gradient-based methods and embedding-based methods, as well as some preliminary explorations to generalize them to LLMs. We then analyze the challenges associated with generalizing the above strategies to LLMs with unique transformer structures and unprecedented numbers of parameters. Finally, we discuss the insights to address the challenges, as well as open challenges worthy of further investigation.

\subsection{Literature Review of Sample-based Explanation}

% Sample-based explanation
% Should we consider in-context learning? Few shot examples are provided in the context

% We use $\mathcal{X}$ to denote the space of input token sequences, and $\mathcal{Y}$ for the space of discrete labels in classification tasks or the space of token sequences as output in generation tasks. Accordingly, we have a training dataset $\mathcal{D}_{train} = \{z_{i} = (x_{i}, y_{i})\}_{i=1}^{N}$ with $N$ samples drawn from the joint space $\mathcal{X} \times \mathcal{Y}$, on which an LLM model $f_{\theta}$ is trained with pretrained parameters $\hat{\theta} \in \mathbb{R}^{P}$. We also have a test sample $z=(x, y)$ of interest, where we want to explain the generation of $y$ from $x$ based on training samples in $\mathcal{D}_{train}$. The goal of sample-based explanation is to measure the influence of a training sample $z_{i} \in \mathcal{D}_{train}$ or a certain segment within $z_{i}$, such that the generation of LLMs can be well-explained and backed up by the selected training samples. 

\subsubsection{Influence Function-based Methods}

One strategy to quantify the influence of a training sample $z_{i}$ in the dataset $\mathcal{D}_{train}$ to a test sample $z$ is through the influence function \citep{koh2017understanding,han2020explaining}, which measures the change of the prediction loss $\mathcal{L}\left(z, \theta \right)$ for $z$, when the training sample $z_{i}$ is removed from the dataset:
\begin{equation}
\label{eq:influence_func}
\mathcal{I}(z_{i}, z) = -\nabla_\theta \mathcal{L}(z, \hat{\theta})^{\top} \mathbf{H}_{\hat{\theta}}^{-1} \nabla_\theta \mathcal{L}(z_{i}, \hat{\theta}),
\end{equation}
where $\nabla_\theta \mathcal{L}(z, \hat{\theta})$ is the gradient of loss $\mathcal{L}$ on $z$, and $\mathbf{H}_{\hat{\theta}}\stackrel{\text {def}}{=} \frac{1}{N} \sum_{i=1}^N \nabla_\theta^2 \mathcal{L}(z_i, \hat{\theta})$ is the Hessian matrix.  To improve efficiency, \citet{koh2017understanding} adopt an iterative approximation process to calculate the Hessian-Vector Product (HVP) in Eq. (\ref{eq:influence_func}), where the memory complexity can be reduced to $\mathcal{O}(P)$ and time complexity to $\mathcal{O}(NPr)$ ($r$ is the number of iterations). To further reduce the complexity, \cite{pruthi2020estimating} propose TracIn, which measures the influence by calculating the total reduction of loss whenever $z_{i}$ is included in the minibatch during model training as follows:
\begin{equation}
\label{eq:tracin}
\mathcal{I}_{\operatorname{TracIn}}\left(z_{i}, z\right)=\sum_{t: z_{i} \in \mathcal{B}_{t}} \mathcal{L}\left(z, \theta_t\right)-\mathcal{L}\left(z, \theta_{t+1}\right) \approx 
\frac{1}{|B_t|} \sum_{t: z_{i} \in B_t} \eta_t \nabla_{\theta} \mathcal{L}\left(z_{i}, \theta_t\right) \cdot \nabla_{\theta} \mathcal{L}\left(z, \theta_t\right),
\end{equation}
where $\mathcal{B}_{t}$ is the $t$-th mini-batch, $\theta_t$ is the parameter at the $t$-th step, $\eta_t$ is the step size. TracIn only leverages gradients, which substantially improves the efficiency. In addition, \citet{schioppa2022scaling} propose to use Alnordi iteration \citep{arnoldi1951principle} to find the dominant eigenvalues and eigenvectors of $\mathbf{H}_{\hat{\theta}}$ on randomly sampled subsets $\mathcal{D}_{sub}$, with $|\mathcal{D}_{sub}| \ll |\mathcal{D}_{train}|$, where the diagonalized Hessian can be cheaply cached and inverted. Observing that finding the most influential training sample on $z$ needs to iterate overall $N$ training samples, \cite{guo2021fastif} propose to use fast KNN to pre-filter a small subset of influence-worthy data points from $\mathcal{D}_{train}$ as candidates to explain small pretrained language models, whereas \cite{han2022orca} propose to find a small subset $\mathcal{D}_{sub} \subset \mathcal{D}_{train}$ whose gradient is the most similar to the downstream task examples. \cite{grosse2023studying} propose to use the Eigenvalue-corrected Kronecker-Factored Approximate Curvature (EK-FAC) approximation to scale influence functions to 52B LLMs. For adaptation, only influences mediated by the feed-forward networks (FFNs) are considered. Based on the assumption that weights from different FFN layers are independent, the EK-FAC approximated influence is finally formulated as the sum of influences mediated by each layer as follows:
\begin{equation}
\label{eq:ekfac}
\mathcal{I}_{\text{EKFAC}}(z_{i}, z) =  \sum_{l} \nabla_{\theta^{(l)}} \mathcal{L}(z, \hat{\theta})^{\top} (\hat{\mathbf{G}}_{\hat{\theta}^{(l)}} + \lambda^{(l)} \mathbf{I})^{-1} \nabla_{\theta^{(l)}} \mathcal{L}(z_{i}, \hat{\theta}),
\end{equation}
where $\theta^{(l)}$ denotes the weights of the $l$-th MLP layer, and $\hat{\mathbf{G}}_{\hat{\theta}^{(l)}}$ is the EK-FAC approximated Gauss-Newton Hessian for $\theta^{(l)}$. Since the inversion of $L$ small ${K_{l} \times K_{l}}$ matrices (i.e., $\mathcal{O}(L \times K_{l}^{3})$) is substantially more efficient than the inversion of a large $L{K_{l} \times LK_{l}}$ matrix (i.e., $\mathcal{O}{((LK_{l}})^{3})$), $\mathcal{I}_{\text{EKFAC}}$ can be adaptable to very large models. 

Recently, influence functions have also been used to explain and improve the in-context learning ability of LLMs. Based on the finding~ \cite{von2023transformers} that during in-context learning, LLMs can be viewed as implicitly "learning" an internal kernelized least square surrogate objective on the few-shot examples included in the prompt, \citet{zhou2024detail} propose to use this hypothesized objective as the loss function to calculate the influence of each in-context sample on the LLM generation, and based on which select and re-rank the samples. They show that the in-context learning ability of LLMs can be improved after the sample selection and demonstration order re-ranking.
% \citep{lampinen2022can}

\subsubsection{Embedding-based Methods}

Another strategy for sample-based explanation leverages hidden representations within the transformer architecture, which is recognized for encoding high-level semantics from textual data, to calculate the semantic similarity between $z$ and $z_{i}$. The similarity can also be used to measure the influence of $z_{i}$ on $z$ as explanations \citep{rajani2019explain}. Specifically, \cite{akyurek2022tracing} propose to represent the training sample $z_{i}$ and test sample $z$ by concatenating the input and output as $z^{cat}_{i} = [x_{i} || y_{i}]$, $z^{cat} = [x || y]$. The concatenation is feasible for generation tasks where the output $y$ lies in the same token sequence space as the input prompt $x$. The similarity between $z_{i}$ and $z$ can then be calculated as:
\begin{equation}
\label{eq:emb}
\mathcal{I}_{\text{emb}}\left(z_{i}, z\right)= \frac{f^{(l)}_{\hat{\theta}}(z^{cat}_{i})^{\top} \cdot f^{(l)}_{\hat{\theta}}\left(z^{cat}\right)}{\left\|f^{(l)}_{\hat{\theta}}(z^{cat}_{i})^{\top}\right\|\left\|f^{(l)}_{\hat{\theta}}\left(z^{cat}\right)\right\|} ,
\end{equation}
where $f^{(l)}_{\hat{\theta}}$ is the sub-network that outputs the $l$-th layer intermediate activation of $f_{\hat{\theta}}$. The Eq. (\ref{eq:emb}) has a similar form as the vanilla influence function defined in Eq. (\ref{eq:influence_func}) as well as its TracIn alternative defined in Eq. (\ref{eq:tracin}), which assigns a score $\mathcal{I}$ for the explainee $z$ for each training sample $z_i$ in the dataset $\mathcal{D}_{train}$ as the explanation confidence of the sample $z_{i}$. 

% Compared with the influence function methods introduced in the previous part, embedding-based methods are computationally efficient, as for each explainee $z$, the explanation score from a training sample $z_{i}$ requires only one forward pass of the transformer network. In addition, the calculation can be easily paralleled for different training samples. However, the disadvantage is also evident: These methods lack a theoretical foundation and may fail to identify important training samples that may not be semantically similar to the test sample. Consider the following toy example: Training samples $z_{i}$ = (``1+1='', ``2'')  and $z_{j}$ = (``2+2='', ``4'') make the LLM gain the ability to conduct arithmetic calculation, which explains why prompting the model with $x$ = ``100+100'' gives the results $y$=``200''. However, the embeddings between the test sample $z$ and the two training samples $z_{i}$ and $z_{j}$ can be very different when calculated via Eq. (\ref{eq:emb})~\citep{akyurek2022tracing}. Therefore, embedding-based methods may not be able to faithfully find the training samples where the explanations require generalization ability beyond semantic similarity.

\subsection{Case Study: EK-FAC-based Influence Estimation}

This case study implements the EK-FAC-based influence function~\cite{grosse2023studying} and verifies its scalability and effectiveness on LLMs with billions of parameters, including GPT2-1.5B~\citep{radford2019language}, LLaMA2-7B~\citep{touvron2023llama2}, Mistral-7B~\citep{jiang2023mistral}, and LLaMA2-13B.

\subsubsection{Experimental Design}

We use the SciFact dataset \citep{wadden2020fact} as the corpora, which contains the abstract of 5,183 papers from basic science and medicine. The LLMs are obtained by finetuning the pretrained LLMs for 20,000 iterations, where AdamW is used as the optimizer \citep{loshchilov2018decoupled}. Then, we use 500 samples from the corpora to estimate the \textbf{\textit{(i)}} uncentered covariance matrices of the activations and pre-activation pseudo-gradients $\mathbf{Q}^{(l)}_{A}$, $\mathbf{Q}^{(l)}_{S}$, and \textbf{\textit{(ii)}} the variances of the projected pseudo-gradient $\boldsymbol{\Lambda}^{(l)}$ for each selected dense layer $l$, and cache them on the hard disk (details see Eqs. (16) and (20) in \cite{grosse2023studying}). We select the \texttt{c\_fc} layer for GPT2-1.5B, and \texttt{gate\_proj} layer for LLaMA2-7B, Mistral-7B, and LLaMA2-13B. For evaluation, we randomly select 200 samples to construct the test set, which we name SciFact-Inf. Specifically, for the $j$-th sample $z_{j} = x_{j}$, we use the first three sentences in $x_{j}$, i.e., $\hat{x}_{j}$, to generate a completion $\hat{y}_{j}$ with the finetuned LLM. Ideally, the $j$-th training sample $z_{j}$ itself should be the most influential sample w.r.t. the generation of $\hat{y}_{j}$. 

% For evaluating each test sample $\hat{z}_{j}$, we first calculate the EK-FAC approximated HVP part of the influence $\mathcal{I}_{\text{EKFAC}}(z_{i}, \hat{z}_{j})$, i.e., $\sum_{l} \nabla_{\theta^{(l)}} \mathcal{L}(\hat{z}_{j}, \hat{\theta})^{\top} (\hat{\mathbf{G}}_{\hat{\theta}^{(l)}} + \lambda^{(l)} \mathbf{I})^{-1}$, which is shared for all training samples $z_{i}$. Specifically, we record the layer-wise gradient $\nabla_{\theta^{(l)}} \mathcal{L}(\hat{z}_{j}, \hat{\theta})$ and calculate the HVP with the cached $\mathbf{Q}_{A}^{(l)}$, $\mathbf{Q}_{S}^{(l)}$ as Eq. (21) in \cite{grosse2023studying}. We then go through candidate training samples (1 positive and 99 negative), calculate the gradient $\nabla_{\theta^{(l)}} \mathcal{L}(z_{i}, \hat{\theta})$, and take inner-product with the approximate HVP as the layer-wise influence. Finally, the layer-wise influences are summed up as Eq. (\ref{eq:ekfac}) as the total influence $\mathcal{I}_{\text{EKFAC}}(z_{i}, \hat{z}_{j})$. We rank the influence and calculate the top-$K$ hit rate of the positive training sample. 

\subsubsection{Results and Analysis}

\begin{table}[t]
\small
\centering
% \vspace{-0.5cm}
\caption{Effectiveness of EK-FAC approximated influence function on the SciFact-Inf dataset. Time (Pre.) stands for the time for precomputing the $\mathbf{Q}_{A}$, $\mathbf{Q}_{S}$, and $\boldsymbol{\Lambda}$. Time (Inf.) stands for time for calculating the influence of 100 training samples per test sample.}
\label{tab:influence}
\begin{tabular}{@{}rlcccc@{}}
\toprule
Strategy & LLM              & Recall@5 & Recall@10 & Time (Pre.)  & Time (Inf.) \\ 
\midrule 
\multirow{4}{*}{Inf. Func.} & GPT2-1.5B        & 0.6368  & 0.7363 & 0h 27min & 0min 28sec \\
& Mistral-7B       & 0.6418  & 0.6866 & 2h 05min & 1min 47sec\\
& LLaMA2-7B        & 0.8063  & 0.8308 & 1h 37min & 1min 34sec\\
& LLaMA2-13B       & 0.7811  & 0.8940 & 3h 11min & 3min 08sec\\
\bottomrule
\end{tabular}
\end{table}The results are summarized in Table~\ref{tab:influence}. We observe that the EK-FAC approximated influence function achieves good accuracy in finding the training sample with greatest influence on the generation of LLMs. In addition, we find that the main computational bottleneck in calculating the EK-FAC-based influence is to estimate the covariances $\mathbf{Q}^{(l)}_{A}$, $\mathbf{Q}^{(l)}_{S}$ and variance $\boldsymbol{\Lambda}^{(l)}$, which takes hours when 500 training samples are used for the estimation. However, after that, it is relatively cheap to calculate the influence training samples for each test sample.

\subsection{Challenges}

Overall, explaining the generation of LLMs by tracing back to the training samples is still an emerging area. Open questions need to be addressed to further advance the field. 

\subsubsection{Strong Assumptions for Scalability} The unprecedented number of parameters in modern LLMs causes severe scalability issues for sample-based explanation strategies. This is especially evident for the gradient-based methods, as the HVP in Eq. (\ref{eq:influence_func}) induces both high computational and space complexity. To address the bottleneck, strong assumptions are usually required to make it feasible for large models. For example, TracIn \citep{pruthi2020estimating} simplifies the second-order term in Eq. (\ref{eq:influence_func}) via first-order approximation. \cite{schioppa2022scaling} assume the Hessian to be low rank. \cite{grosse2023studying} that assume that the weights from different layers of the LLMs are independent, as well as the tokens in different steps, such that EK-FAC can be appropriately applied to approximate the influence function. From the above analysis, we can find that while the method from \cite{grosse2023studying} has the best scalability, it also has the strongest assumption, which may fail to hold in practice. Therefore, how to improve the scalability with weak assumptions needs to be investigated in the future.

% \subsubsection{Explainability v.s. Understandability}

% Despite the advantage of influence/embedding similarity to provide a qualitative measurement of a specific training sample as the \textbf{explanation} for LLM generation, the \textbf{understandability} of the identified sample can still be weak, where the connection between the selected training samples and the generation may not be understandable to human beings. Specifically, \cite{grosse2023studying} cautions that the sign of influence score of the training tokens may be difficult for humans to connect to the positive or negative influence on the generation results. This severely jeopardizes the usability of the identified training samples. In addition, \cite{grosse2023studying} also found that, since LLMs are usually not trained to the minimum to avoid overfitting (and due to overparameterization, the number of local minimums may be large), the connection between influence defined in Eq. (\ref{eq:influence_func}) with the counterfactual loss of removing the sample $z_{i}$ at $z$ is also weak. For the embedding-based methods, since most LLM models are black box transformer models, the similarity of embeddings can also be hard to interpret by human beings; therefore, it is imperative to improve the interpretability of the identified training samples, such that tracing back becomes more meaningful.

\subsubsection{LLM-Oriented Sample-based Explanations}

Finally, we observed that both gradient-based and embedding-based methods are loosely connected to the LLM, as well as the backbone transformer networks. For example, algorithms like TracIn \citep{pruthi2020estimating} are designed to scale up influence functions to large models, which are not specific for LLMs. Similarly, the embedding-based method proposed in \cite{akyurek2022tracing} is applicable to most machine learning models with latent representations. \cite{grosse2023studying} considers the specialty of LLMs by utilizing the knowledge neuron assumption of the backbone transformers \citep{wang2023knowledge} to simplify the influence function, where the weights considered are constrained to the MLP layers, which may not fully utilize the property of transformers. Therefore, how to further utilize the property of the LLM and the backbone transformer to design LLM-tailored sample-based influence/similarity (either to reduce the computational/space overhead or to improve the explanation quality) is highly promising for future work.

\section{Explainability for Trustworthy LLMs and Human Alignment}\label{sec:diagnosis}

In previous sections, we explore the use of explanation techniques for assessing and improving the performance of LLMs.
In this section, we shift the focus towards examining LLM trustworthiness. 
As LLMs are increasingly integrated into high-stakes areas like healthcare, finance, and legal advice, it is crucial that their responses not only are accurate but also \textit{align} with human ethical standards and safety protocols~\citep{liu2023trustworthy,li2023survey,hu2023decipherpref}. Herein, we examine how explanation techniques, discussed in the previous sections, can be instrumental in assessing LLMs across key aspects of trustworthiness like security, privacy, fairness, toxicity, and honesty.
It is worth noting that while explainability itself is an aspect of trustworthiness, it holds the promise of serving as a foundational tool for addressing other trustworthiness concerns.

\subsection{Security}
LLMs are known to be vulnerable to attacks and exploitation, such as spreading misinformation, and poisoning training data~\citep{derner2023security}. 
For enhanced safety, LLMs are designed to reject prompts that may result in harmful contents. However, jailbreak techniques can circumvent these restriction measures and manipulate LLMs into producing malicious contents. 
Malevolent users can craft special prompts that compel LLMs to prioritize instruction following over rejections~\citep{liu2023jailbreaking,li2023multi}. 
For example, through Prefix Injection, attackers can use out-of-distribution prompt prefixes that are less likely to be rejected~\citep{walker2022dan, wei2023jailbroken}. Another approach, called Refuse Suppression, involves directing or persuading models to ignore established safety protocols~\citep{wei2023jailbroken,zeng2024johnny}, where the instruction following ability is then employed to perform the attack. 

\tikzstyle{leaf1}=[
my-box, 
minimum height=1.5em,
fill=blue!10, 
text=black,
align=left,
font=\tiny,
inner xsep=2pt,
inner ysep=3pt,
text width=11em,
]

\begin{figure}[t]
\vspace{-0.2cm}
\centering
\resizebox{0.8\textwidth}{!}{
\begin{forest}
forked edges,
for tree={
    grow=east,
    reversed=true,
    anchor=base west,
    parent anchor=east,
    child anchor=west,
    rectangle,
    draw,
    font=\tiny,
    rounded corners,
    align=center,
    minimum width=2em,
    edge+={darkgray, line width=1pt},
    s sep=4pt,
    inner xsep=3pt,
    inner ysep=2pt,
    ver/.style={rotate=90, child anchor=north, parent anchor=south, anchor=center, text centered},
},
[Trustworthy LLMs \& Human Alignment,ver,fill=gray!20,text width=10em,text centered
  [Security,ver,fill=orange!40,text centered,text width=2.5em
      [Adversarial Attack,fill=orange!10,text centered,text width=6.5em
          [{\cite{liu2023jailbreaking}, \cite{li2023multi}, \cite{wei2023jailbroken}, \cite{zeng2024johnny}, \cite{walker2022dan}, \cite{liu2021adversarial}, \cite{jain2023mechanistically}.}, leaf1, align=left]
        ]
      [Adversarial Defense,fill=orange!10,text centered,text width=6.5em
      [{\cite{wu2025interpreting}, \cite{yu2024robust}, \cite{zheng2024prompt}, \cite{yousefpour2025representation}.}, leaf1, align=left]
      ]
  ]
  [Privacy,ver,fill=orange!40,text width=2.5em,text centered
   [Memorization Attack,fill=orange!10,text centered,text width=6.5em
          [{\cite{nasr2023scalable}, \cite{li2023multi}, \cite{fu2024think}, \cite{bai2024special}.}, leaf1, align=left]
        ]
    [Knowledge-based Defense,fill=orange!10,text centered,text width=6.5em
          [{\cite{meng2022locating}, \cite{dai2022knowledge}, \cite{hase2024does}, \cite{yin2024benchmarking}, \cite{suri2025mitigating}.}, leaf1, align=left]
        ]
    ]
  [Fairness,ver,fill=orange!40,text width=2.5em,text centered
      [Bias Detection,fill=orange!10,text centered,text width=6.5em
        [{\cite{adebayo2023quantifying}, \cite{ma2023deciphering}, \cite{zou2023representation}, \cite{qian2024dean}.}, leaf1, align=left]
      ]
      [Bias Mitigation,fill=orange!10,text centered,text width=6.5em
       [{\cite{rakshit2024prejudice}, \cite{ma2023deciphering}, \cite{yu2023unlearning}, \cite{thakur2023language}.}, leaf1, align=left]
      ]
  ]
  [Toxicity,ver,fill=orange!40,text width=2.5em,text centered
    [Toxic Content Detection,fill=orange!10,text centered,text width=6.5em
          [{\cite{balestriero2023characterizing}, \cite{wang2024detoxifying}, \cite{lee2024mechanistic}, \cite{li2024open}.}, leaf1, align=left]
      ]
    [Toxic Content Mitigation,fill=orange!10,text centered,text width=6.5em
      [{\cite{lee2024mechanistic}, \cite{liu2024intrinsic}, \cite{leong2023self}.}, leaf1, align=left]
      ]
  ]
  [Truthfulness,ver,fill=orange!40,text width=2.5em,text centered
      [Hallucination Detection,fill=orange!10,text centered,text width=6.5em
          [{\cite{azaria2023internal}, \cite{campbell2023localizing}, \cite{duan2024llms}, \cite{li2024inference}, \cite{marks2023geometry}, \cite{xu2023understanding}, \cite{heo2024llms}, \cite{park2025steer}.}, leaf1, align=left]
      ]
    [Hallucination Mitigation,fill=orange!10,text centered,text width=6.5em
        [{\cite{li2024inference}, \cite{chencontext}, \cite{liu2024intrinsic}, \cite{yu2024mechanistic}, \cite{huang2024dishonesty}, \cite{zhang2024seeing}, \cite{he2024cracking}.}, leaf1, align=left]
      ]
  ]
]
\end{forest}
}
\caption{Taxonomy of explanation techniques for trustworthiness aspects of LLMs.}
\label{fig:taxonomy_trustworthiness}
\vspace{-0.4cm}
\end{figure}

% Existing attacking strategies on LLMs heavily rely on prompt engineering, resulting in a low attack success rate and significant time cost~\citep{li2024open}.
By engineering latent representations of LLMs, explanation methods provide a viable way to discover the potential vulnerabilities of LLMs by simulating advanced attacks~\citep{liu2021adversarial}. 
For example, a recent work extracts ``safety patterns'' via explaining the latent space. These patterns are captured from the activation differences between malicious and benign queries. They reflect the internal protection mechanisms within LLMs. Circumventing these patterns leads to novel attacks, which helps exploring potential vulnerabilities of LLMs~\citep{li2024open}. 
Besides, a deeper understanding of fine-tuning can shed light on the reliability of existing safety measures. In particular, \citet{jain2023mechanistically} apply network pruning, attention maps, and probing classifiers to track the changes of model capabilities from pre-training to fine-tuning. These tools are helpful in demonstrating that the capabilities gained during fine-tuning can be easily removed through fine-tuning on other unrelated tasks. This finding casts doubt on the robustness of current safety alignments in LLMs.
Finally, by probing LLM representations, \citet{zheng2024prompt} investigates the impact of safety prompts that motivates a novel safety prompt optimization method to safeguard LLMs.

\subsection{Privacy}
Recent studies have revealed that LLMs such as ChatGPT can leak extensive amounts of training data through divergence attacks~\citep{nasr2023scalable}. These attacks utilize specially crafted prompts to lead the model away from its standard chatbot-style generation. The risk of private data exposure through such means poses a serious challenge to the development of ethically responsible models. This issue is compounded by strategies similar to jailbreak attacks, where misalignment is exploited to induce LLMs into operating in an unconventional ``developer mode'' via out-of-distribution prompts~\citep{li2023multi}.

Enhancing LLM privacy involves two strategic approaches: (1) preventing LLMs from memorizing sensitive data, and (2) establishing safeguards against the release of sensitive information.
The second strategy involves employing techniques from jailbreak defenses, treating prompts that solicit private information as potentially malicious.
The first approach requires identifying specific knowledge within models, which is traditionally achieved with prompt engineering. However, prompt engineering faces limitations due to LLMs' sensitivity to the phrasing of prompts.
Explanatory techniques can serve as an auxiliary tool to confirm whether LLMs have internalized certain knowledge.
For instance, factual knowledge is localized via explaining the relation between factual knowledge and neuron activations~\citep{meng2022locating,dai2022knowledge,hase2024does}. 
In addition, \cite{yin2024benchmarking} recently proposes the concept of ``knowledge boundary'' and develops a gradient-based method to explore whether LLMs master certain knowledge independent of the input prompt.

\subsection{Fairness}
The widespread applications of LLMs also bring concerns about exacerbating bias issues in society~\citep{gallegos2023bias}. For example, in a gender stereotype case, ``[He] is a doctor'' is much more likely than ``[She] is a doctor''. In this subsection, we focus on fairness issues related to race, gender, and age~\citep{li2023prompt}.
Prompt engineering is a major method to quantify fairness issues within LLMs~\citep{adebayo2023quantifying}. Interpretation complements this method by unraveling the mechanisms through which biases are embedded into LLMs. 
One direction is examining biased attention heads.~\citet{ma2023deciphering} unveil that approximately 15\% to 30\% of attention heads linking to specific stereotypes through probing attention heads.
Furthermore, recent work has placed LLM representations under scrutiny~\citep{zou2023representation}. Typically, representations of crafted templates showing specific biases are examined to derive a vector that identifies a certain bias.

To achieve fair model outputs, a diverse range of mitigation techniques have been proposed. One stream of work proposes to debias LLMs at the embedding level. For example, one work alters biased embeddings with minimal alterations to make them orthogonal to neutral embeddings~\citep{rakshit2024prejudice}. Additionally, some studies remove biases at the level of attention heads. \citet{ma2023deciphering} address this by pruning biased attention heads. Beyond modifying embeddings and pruning attention heads, another strategy targets a specific group of neurons known to propagate biases. It unlearns the biases by retraining weight vectors for these neurons~\citep{yu2023unlearning}. Besides, bias mitigation can also be approached from a data-centric perspective using the most biased training samples, then modifying them to fine-tune the model~\citep{thakur2023language}.

\subsection{Toxicity}
Toxicity is another form of harmful content that LLMs may produce. This issue arises from training documents containing elements of toxicity that can hardly be fully eliminated.
Toxicity can be identified by interpreting LLM components like the feed-forward layers and attention heads. For instance, recent work reveals how toxicity is represented within LLMs by identifying multiple vectors and relevant dimensions promoting toxicity within the MLP layers~\citep{lee2024mechanistic}. 
Furthermore, the exploration of geometric structures in per-layer representations offers another way to detect toxicity. \cite{balestriero2023characterizing} extract input features from MLPs to describe the domain of prompts and classifying toxic remarks.

Aforementioned insights also shed light on mitigation strategies. Motivated by the finding that toxicity can be reduced by manipulating relevant vectors, \citet{lee2024mechanistic} utilizes paired toxic and non-toxic samples to fine-tune models so that non-toxic content is promoted. By examining the changes in the parameter matrices during the fine-tuning process, it substantiates that even minor adjustments to these critical vectors can reduce toxicity. Built on the observation that LLMs' representations are updated by outputs from attention layers~\citep{elhage2021mathematical}, another work attempts to reduce toxicity by identifying the ``toxicity direction'' and then adjusting representations in the opposite direction~\citep{leong2023self}.

\vspace{-0.1cm}
\subsection{Truthfulness}
LLMs tend to confidently produce false statements that fall into two main categories: 1) statements that contradict learned knowledge, also known as \textit{dishonesty}; 2) statements that are factually incorrect and fabricated by models, that is \textit{hallucination}. In the following, we look into explainability tools that aim to understand the above two behaviors. 

\vspace{-0.1cm}
\subsubsection{Dishonesty}
Dishonesty of LLMs describes models' ability to produce false statements compared to their learned information. Studies have been undertaken to understand how and why it happens. One notable work distinguishes dishonesty by training a classifier to predict the accuracy of statements, on top of activations from true/false statements~\citep{azaria2023internal}. The classifier reaches an accuracy range between 60\% and 80\%~\citep{azaria2023internal}. Furthermore, ~\citet{campbell2023localizing} localize dishonesty behaviors at the level of attention heads. Despite adopting above probing-based method, it also employs activation patching to substitute lying activations with honesty ones. Both approaches have witnessed the importance of layer 23$-$29 in flipping dishonesty behaviors. Besides, another popular method tries to study the geometric structure of true/false statements\citep{marks2023geometry}. A linear structure and the truth directions are derived to mitigate the dishonest behaviors.

\vspace{-0.1cm}
\subsubsection{Hallucinations}
Hallucinations in LLMs can arise due to poor data quality and the lack of explicit knowledge~\citep{xu2024hallucination}. However, whether LLMs are aware of their hallucination behaviors remains unknown. Recent work investigates this question by examining models' hidden representation space~\citep{duan2024llms}. It defines an ``awareness'' score to quantify the uncertainty of LLMs' answers, finding that adversarially induced hallucination can increase models' awareness. 
Additionally, \citet{li2024inference} identifies differences between models' output and their inner activations as a potential source of hallucination. By training two orthogonal probes to mitigate hallucinations, they reveal that ``truth'' might exist in a subspace instead of a single direction~\citep{li2024inference}. 
Another work investigates the source of hallucination by analyzing patterns of source token contributions through perturbations~\citep{xu2023understanding}. They find that hallucinations may stem from the model's excessive dependence on a restricted set of source tokens, and the static distribution of source token contributions.

% Building on the above insights into LLM hallucinations, \citet{duan2024llms} apply PCA to derive the direction of the correct answer's final hidden state, and enhance the hidden representations with this direction to reduce hallucinations. In contrast,~\citet{li2024inference} adopts a different approach, by intervening on top-$K$ specialized attention heads, while minimizing the influence of the rest attention heads within models. Different from PCA that identifies a single principle direction, this work adopts two distinct techniques to find multiple directions of intervention. First, they use orthogonal vectors of each probe's hyperplane, which is similar to PCA. Second, they leverage vectors that connect the mean of the true and false distributions~\citep{li2024inference}. The vectors derived from mean shift has been demonstrated more effective than those from probe classifiers, which presents another feasible strategy for identifying directions of truth.

\vspace{-0.1cm}
\subsection{Challenges}
We discuss challenges in employing explanations to improve models' trustworthiness and enhance alignment: 1) limitations of existing evaluation techniques, and 2) shortcomings of mitigation strategies based on explanations.

\vspace{-0.1cm}
\subsubsection{Challenges of Existing Evaluation Methods}

Current detection methods primarily focus on layers, attention heads, and representations. However, we still lack robust understanding of how a certain trustworthiness concept (e.g., fairness, harmlessness) is comprehended by LLMs. Furthermore, we lack general strategies to identify this knowledge. 
Existing methods often rely on building an evaluation dataset, but it is difficult to guarantee the its comprehensiveness. Meanwhile, the definition of what constitutes a trustworthiness concept often varies among different stakeholders and contexts, complicating the evaluation process.
% For instance, to identify gender biases, attention heads might be examined individually rather than adopting a general approach across models~\citep{li2024inference}. 
Moreover, existing localization approaches rely on probing classifiers or casual scrubbing, which might not be reliable, since the performance of probing classifiers depends on the comprehensiveness of pre-defined biases, while casual scrubbing usually introduces new variables that complicate the analysis.

\subsubsection{Challenges of Mitigation Strategies} Currently, mitigation methods for LLMs are typically developed based on explanations. Existing explanations are implemented using techniques from mechanistic interpretability and representation engineering~\citep{zhao2024opening}.
However, neither stream of methods can fully address the issues. For example, the identified directions from representation engineering and patched activations from mechanistic interpretability can only mitigate issues to a certain extent. Moreover, the changes to either representations or activations could also influence other aspects of models' capabilities, which we are yet unable to evaluate.

\section{Effective LLM Inference via Explainable Prompting}\label{sec:prompt_cot}
A key distinction between LLMs and traditional machine learning models lies in the LLMs' ability to accept flexibly manipulated input data, namely prompts, during model inference~\citep{liu2023pre}.
%LLMs generally give precedence to the information presented in these prompts when generating outputs.
To mitigate the opacity issue in LLM predictions, we can enhance prompts with \textit{explainable content}. %, which is then prioritized over the LLMs' inherent and implicit knowledge.
%These enriched prompts can include domain-specific insights, contextual information, or a step-by-step reasoning chain. 
In response, LLMs can reveal their decision-making processes during inference, which improves the explainability and even rationality of their behaviors.
\vspace{-5pt}

\subsection{Explainable Prompting Paradigms and Effects}
% \subsubsection{Explanation-Guided Prompting.}
% The Chain of Thought (CoT) approach significantly enhances LLMs in tackling complex tasks~\citep{wei2022chain}. While LLMs are adept at generating human-like responses, they often lack transparency in their reasoning processes. This limitation makes it difficult for users to assess the credibility of the responses, especially for those requiring detailed reasoning.
% To bridge this gap, recent efforts have incorporated in-context learning with human-crafted explanations directly into prompts~\citep{wei2022chain, huang2023chain,yao2023tree,besta2023graph}. 
To better understand and control how LLMs generate outputs, many researchers directly prompt LLMs to provide the rationales along with their predictions~\citep{wei2022chain, huang2023chain,yao2023tree,besta2023graph,kang2024mindstar}. 
\textbf{Chain-of-Thought} (CoT) prompting~\citep{wei2022chain} stands out by employing explicit reasoning to guide the generation process. Formally, we define the language model as $f_{\theta}$, and input prompt as $X = \{x_1, y_1, x_2, y_2,..., x_n\}$, where $x_1, y_1, x_2, y_2,..., x_{n-1}, y_{n-1}$ denote the example question-response pairs for in-context learning, and $x_n$ is the actual question. In a standard prompting scenario, we have the model output as $y_n = \arg\max_{Y} p_{\theta}(Y | x_1, y_1, x_2, y_2,..., x_n)$. However, this approach does not provide insight into how the answer $y_n$ is generated. Therefore, CoT proposes including human-crafted explanations $e_i$ for the $i$-th in-context example, resulting in a modified input format $X = \{x_1, e_1, y_1, x_2, e_2, y_2,..., x_n\}$. Given the input, the model will output not only prediction $y_n$ but along with an explanation $e_n$:
%\begin{equation}
$
    e_n, y_n = \arg\max_{Y} p_{\theta}(Y |x_1, e_1, y_1, x_2, e_2, y_2,..., x_n) .
$
%\end{equation}
%Besides allowing for a more transparent and understandable interaction with LLMs, the CoT approach is also practically useful as it augments LLMs' functionality by opening a window for users to control the models' thought processes. 
Specifically, the usefulness of CoT methods lies in several key aspects:
\begin{itemize}[leftmargin=*, topsep=0pt, itemsep=0pt]
    \item \textbf{Reducing Errors in Reasoning:} By breaking down complex problems into a series of smaller tasks, CoT reduces errors in logic-oriented tasks, leading to a more precise resolution of intricate problems~\citep{wei2022chain, qin2023chatgpt, zhang2023multimodal, wang2024chain}.
    \item \textbf{Providing Adjustable Intermediate Steps:} CoT enables the outlining of traceable intermediate steps within the problem-solving process. This feature enables users to trace the model’s thought process from inception to conclusion, and to adjust the prompts if undesirable model behaviors are observed~\citep{lyu2023faithful, wang2023knowledge}. 
    \item \textbf{Facilitating Knowledge Distillation:}  The step-by-step reasoning derived from larger LLMs can serve as a specialized fine-tuning dataset for smaller LLMs. It allows smaller models to learn complex problem solving by following explanations, effectively teaching them to tackle intricate questions with enhanced reasoning capabilities~\citep{magister2022teaching}.
\end{itemize}

%\subsection{Extended Methods of Explainable Prompting}
%\noindent \textbf{Extended Methods of Explainable Prompting.} 

% \textbf{Enhanced CoT Prompting.}
\noindent In addition, techniques beyond simple CoT have been developed to broaden the range of reasoning paths available to LLMs~\citep{zelikman2022star,yao2023tree, besta2023graph, yao2023beyond, chen2024unlocking,zelikman2024quiet}.  %dhuliawala2023chain, lyu2023faithful,parvez2024evidence,
%We introduce several notable examples below.
%\textbf{Tree-of-Thoughts (ToT).} 
For example, \textbf{Tree-of-Thoughts (ToT)}~\cite{yao2023tree} advances beyond the traditional linear chain-of-thought reasoning, offering a more versatile structure that allows models to navigate through multiple reasoning paths.
%ToT makes the reasoning process of LLMs more interpretable by closely aligning it with human thought processes, as humans naturally consider multiple options and possible outcomes in both forward planning and retrospective analysis to reach conclusions~\citep{sloman1996empirical, stanovich1999rational}. 
%This capability enhances the capacity of LLMs to tackle complex challenges that require the ability to consider and reevaluate different strategies, such as devising game strategies or generating creative content. By simulating the way humans think and make decisions, ToT not only makes their thought process more understandable to human users, but also improves the models' effectiveness in handling complex tasks. 
Similarly, \textbf{Graph of Thoughts (GoT)} \cite{besta2023graph} transforms the outputs into a graph, where information pieces are nodes and their connections are edges, enabling a more intricate and connected form of reasoning paths. %compared to previous methods. 
 By organizing data into nodes (individual concepts or pieces of information) and edges (relationships between these concepts), GoT makes the logical connections within complex systems more understandable~\citep{yao2023beyond}. 
%This graphical representation enables dynamic modification of relationships between concepts, offering a clear visualization of how changing one element affects the others. This is crucial in fields like legal reasoning~\citep{cui2023chatlaw, boche2024mathematical}, scientific research~\citep{ding2023everything, dechoudhury2023benefits}, and policy analysis~\citep{chen2023can}, where the inter-dependencies between various factors can be intricate and subtle. Secondly, GoT enables an assessment of the significance of each node within the graph, providing insights into which pieces of information are most critical to the task. This level of adaptability and clarity makes GoT exceptionally powerful for analyzing and navigating complex information networks.

% \subsubsection{Learning-based Explainable Prompting.}
However, the above prompting-based methods may not always trigger LLMs to provide explanations. 
Thus, many efforts~\cite{cheng2024chainlm,zhang2024chain,pang2024iterative,yang2024weak} have been made to empower LLMs to acquire the inherent ability of providing their reasoning paths.
DeepSeek-R1~\cite{guo2025deepseek} is one of the remarkable checkpoints, where they train LLMs to explain before making final predictions, and only give a positive reward if it provides a correct prediction. Even though there are no human-annotated explanations as ground truth to guide the LLM training, the researchers found that LLM can still learn to provide human-understandable explanations and achieve high accuracy in solving reasoning problems.
To ensure the efficiency and faithfulness of LLM-generated explanations, recent works focus on encouraging LLMs to provide a shorter but necessary explanation~\cite{luo2025o1,hou2025thinkprune,yang2025think}.

%\vspace{-0.6cm}
\subsection{Case Study: Is CoT Really Making LLM Inferences Explainable?}
\subsubsection{Background and Experimental Settings}
Despite the apparent intuitiveness of the CoT prompt design, a critical question remains unanswered: \textit{Does CoT really make LLM inferences explainable?} In other words, can the information provided through CoT faithfully reflect the underlying generation process of LLMs?
We use multi-hop question-answering (QA) as the scenario to investigate this problem.
In QA systems, answering multi-hop questions remains a significant challenge. Instead of leveraging a single information source, multi-hop questions require synthesizing information from multiple pieces or sources of data into a coherent and logical sequence~\citep{tan2023can, kim2023sure, zhong2023mquake}. 
%While LLMs show good performance in single-hop QA tasks~\citep{radford2019language}, their efficacy significantly declines in multi-hop situations~\citep{tan2023can, kim2023sure, zhong2023mquake}. This discrepancy highlights the need for more advanced methods to effectively handle the intricacy of multi-hop reasoning.

\begin{wrapfigure}{r}{0.3\textwidth}
    \centering
    \vspace{-0.4cm}
    \includegraphics[width=0.3\textwidth]{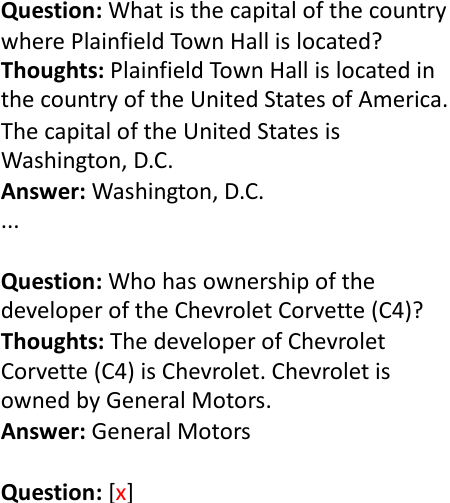}
    \vspace{-0.8cm}
    \caption{CoT Prompts for Multi-hop QA.}
    %\vspace{-0.3cm}
    %\vspace{-0.4cm}
    \vspace{0.1cm}
    \includegraphics[width=0.30\textwidth]{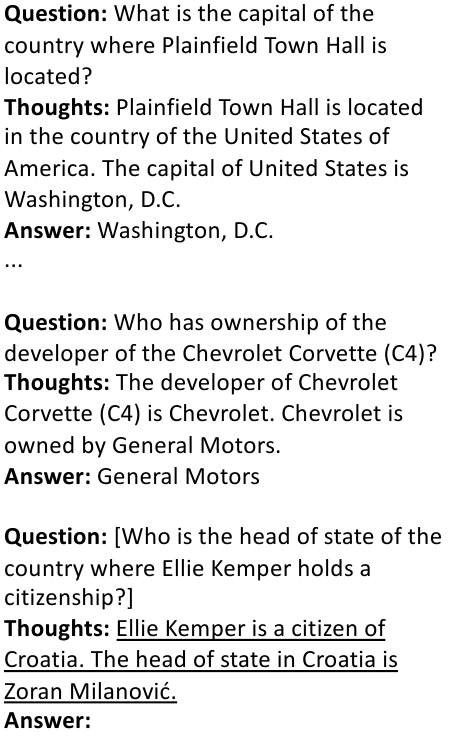}
    \vspace{-0.9cm}
    \caption{CoT Faithfulness for Explanation.}
    \label{pre}
    \vspace{-0.8cm}
\end{wrapfigure}
\textbf{CoT Prompts for Multi-hop QA.} 
To address this, our case study applies the CoT technique with in-context examples. We provide an example as follows, where $[x]$ denotes the question. The ``Thoughts'' following each ``Question'' are step-by-step problem-solving statements for the multi-hop questions.

%The thoughts in the templates align the generation process of LLMs with human cognitive problem-solving patterns. 
%for a sequence of simpler, linked sub-questions which are deconstructed by multi-hop questions. 

% \begin{lstlisting}
% Question: What is the capital of the country where Plainfield Town Hall is located?
% Thoughts: Plainfield Town Hall is located in the country of the United States of America. The capital of the United States is Washington, D.C.
% Answer: Washington, D.C.
% ...

% Question: Who has ownership of the developer of the Chevrolet Corvette (C4)?
% Thoughts: The developer of Chevrolet Corvette (C4) is Chevrolet. Chevrolet is owned by General Motors.
% Answer: General Motors

% Question: [x]
% \end{lstlisting}
% \begin{wrapfigure}{r}{0.30\textwidth}
%     \centering
%      \vspace{-.6cm}
%     \includegraphics[width=0.30\textwidth]{Figures/Template2.pdf}
%     \vspace{-0.9cm}
%     \caption{CoT Faithfulness for Explanation.}
%     \label{pre}
%     \vspace{-10pt}
% \end{wrapfigure}
% \begin{wrapfigure}{r}{0.3\textwidth}
%     \centering
%     \vspace{-0.4cm}
%     \includegraphics[width=0.30\textwidth]{Figures/Template2.pdf}
%     \vspace{-0.9cm}
%     \caption{CoT Faithfulness for Explanation.}
%     \label{pre}
%     \vspace{-0.4cm}
% \end{wrapfigure}
\textbf{CoT Faithfulness for Explanation:} 
To quantitatively measure the faithfulness of CoTs, we select fidelity as the corresponding metrics~\citep{zhao2023explainability, wachter2017counterfactual}:
\begin{flalign*}
   \enspace\enspace\enspace\enspace\enspace\enspace\enspace\enspace Fidelity =\frac{\sum_{i=1}^{N}\left(\mathbbm{1}\left(\hat{\mathbf{y}}_{i}=\mathbf{y}_{i}\right)-\mathbbm{1}\left(\hat{\mathbf{y}}_{i}^{mislead}=\mathbf{y}_{i}\right)\right)}{\sum_{i=1}^{N}\left(\mathbbm{1}\left(\hat{\mathbf{y}}_{i}=\mathbf{y}_{i}\right)\right)}  \times 100\%, &&
\end{flalign*}
where $\mathbf{y}_{i}$ denotes the ground truth label, $\hat{\mathbf{y}}_{i}$ denotes the original model output with CoT, while $\hat{\mathbf{y}}_{i}^{mislead}$ denotes the model output with misleading information inserted. 
In the following, we give an example. Given the target question, the correct step-by-step thoughts should be: ``\textit{Ellie Kemper is a citizen of the United States of America. The president of the United States of America is Joe Biden.}''
To mislead the model, we replace the thoughts with incorrect information (the underlined text) and ask the model to generate a new answer based on incorrect thoughts. If the model still generates the correct answer after the modification, we believe that the CoT information does not faithfully reflect the true process of the answer generation. On the other hand, if it generates an answer corresponding to the incorrect thoughts, then we claim the thoughts are faithful.
% \begin{lstlisting}[escapechar=\%]
% Question: What is the capital of the country where Plainfield Town Hall is located?
% Thoughts: Plainfield Town Hall is located in the country of the United States of America. The capital of United States is Washington, D.C.
% Answer: Washington, D.C.

% ...

% Question: Who has ownership of the developer of the Chevrolet Corvette (C4)?
% Thoughts: The developer of Chevrolet Corvette (C4) is Chevrolet. Chevrolet is owned by General Motors.
% Answer: General Motors

% Question: [Who is the head of state of the country where Ellie Kemper holds a citizenship?]
% Thoughts: %\underline{Ellie Kemper is a citizen of Croatia. The head of state in Croatia is Zoran Milanović.}%
% Answer:  
% \end{lstlisting}

\textbf{Experimental Settings.} 
We evaluate the performance on the MQUAKE-CF dataset~\cite{zhong2023mquake}, which includes 1,000 cases for each $K$-hop questions, $K \in \{2, 3, 4\}$, which totally consists of 3,000 questions. Our evaluation applies various language models, including GPT-2 (1.5B)~\citep{radford2019language}, GPT-J (6B)~\citep{gpt-j} with 6 billion parameters, LLaMA (7B)~\citep{touvron2023open}, Vicuna-v1.5 (7B)~\citep{chiang2023vicuna},  LLaMA2-chat-hf (7B)~\citep{touvron2023llama2},  Falcon~\citep{almazrouei2023falcon} with 7 billion parameters,  Mistral-v0.1 (7B)~\citep{jiang2023mistral}, and  Mistral-Instruct-v0.2 (7B)~\citep{jiang2023mistral}.
These models have demonstrated proficiency in both language generation and comprehension. 

\subsubsection{Experiment Results}
\begin{wraptable}{r}{0.65\textwidth}
%\begin{table*}[h]
\renewcommand\arraystretch{0.8}
\small
\centering
\vspace{-0.4cm}
\caption{CoT Faithfulness Evaluation on MQUAKE-CF.}
\vspace{-0.4cm}
\label{cf2}
%\resizebox{2.00\columnwidth}{!}{
\begin{tabular}{@{}l|ccc|ccc|ccc@{}}
\toprule
Datasets                                   & \multicolumn{9}{c}{MQUAKE-CF}                                                                                                                                                                                                                 \\ \hline
Question Type& \multicolumn{3}{c|}{2-hops}                                                    & \multicolumn{3}{c|}{3-hops}                                                     & \multicolumn{3}{c}{4-hops}                                                   \\ \hline
Edited Instances& \multicolumn{1}{c|}{C} & \multicolumn{1}{c|}{M} & \multicolumn{1}{c|}{F} & \multicolumn{1}{c|}{C} & \multicolumn{1}{c|}{M} & \multicolumn{1}{c|}{F} & \multicolumn{1}{c|}{C} & \multicolumn{1}{c|}{M} & \multicolumn{1}{c}{F} \\ \hline
%Model                                     &                         &                          &                          &                         &                          &                           &                         &                         &                          \\
\multirow{1}{*}{GPT-2 (1.5B)}                     &                         15.9&                          5.2& 67.3\%
&                         8.9&                          2.9& 67.4\%
 &                         8.4&                         1.3&                          84.5\%
\\
                                         % & RAG      &                         &                          &                          &                         &                          &                           &                         &                         &                          \\\hline
\multirow{1}{*}{GPT-J (6B)}                     &                         51.9&                          7.3&                          85.9\%
&                         30.5&                          1.8&                           94.1\%
&                         49.8&                         2.0&                          96.0\%
\\
                                          %& RAG        &                         &                          &                          &                         &                          &                           &                         &                         &                          \\\hline
\multirow{1}{*}{LLaMA (7B)}                    &                         65.1 &                          9.9&84.8\%&                         39.3&                          6.1&                           84.5\%&                         62.9&                         6.0&                          90.5\%\\ \hline
                                          %& RAG       &                         &                          &                          &                         &                          &                           &                         &                         &                          \\ 
\multirow{1}{*}{Vicuna-v1.5 (7B)}                    & 56.3                         &                          21.7& 61.5\%&                          29.7&                           12.7&                            57.3\%&                          53.1&                          16.1&                           69.7\%\\
\multirow{1}{*}{LLaMA2 (7B)}                    & 58.7& 17.0& 71.0\%&                          30.3&                           8.3&                            72.5\%&                          49.1&                          12.0&                           75.6\%\\
\multirow{1}{*}{Falcon (7B)}                    & 61.7&                          24.0& 61.1\%&                          31.6&                           15.0&                            52.6\%&                          48.6&                          23.1&                           52.4\%\\
\multirow{1}{*}{Mistral-v0.1 (7B)}                    & 69.3                         &                          24.0& 65.4\%&                          42.3&                           13.0&                            69.3\%&                          63.2&                          18.4&                           70.8\%\\
\multirow{1}{*}{Mistral-v0.2 (7B) }                    & 56.3                         &                          47.9& 14.8\%&                          37.7&                           22.0&                            41.6\%&                          56.2&                          37.3&                           33.6\%\\
                                          \bottomrule
\multicolumn{10}{l}{{\scriptsize\itshape C: CoT; M: Misleaded CoT; F: Fidelity Score.}}
\end{tabular}
%\end{table*}
\vspace{-0.6cm}
\end{wraptable}
\textbf{Faithfulness Evaluation of CoT.} 
Table~\ref{cf2} illustrates the impact of accurate versus misleading CoTs on the performance of LLMs. The Fidelity metric indicates how faithfully the model's output reflects the reasoning process described in the CoT. Ideally, a high Fidelity score suggests that the model's final response is directly based on the provided CoT, validating it as a faithful explanation of the model's reasoning pathway. However, as we will discuss below, a low Fidelity may not always imply a lack of faithfulness in the model's reasoning, which calls for developing more effective evaluation methods in future research. 

GPT-J and LLaMA exhibit high fidelity scores across different question types, indicating a strong adherence to the given reasoning paths. Conversely, other models show relatively high mislead accuracy scores with lower fidelity scores. In the experiments, we observe that these models usually rely on their own generated thoughts instead of using incorrect information provided in the CoT. Mistral-v0.2, in particular, demonstrates the lowest fidelity scores and highest misleading accuracy scores, suggesting a potential self-defense ability against false information. The lower fidelity scores of later models may be attributed to their improved training processes on more diverse and high-quality datasets, enabling them to develop a better understanding of context and reasoning. As a result, they are more likely to generate their own correct reasoning paths.
%In conclusion, the generated thoughts can be generally viewed as faithful explanations of their output answer. While high fidelity scores generally indicate a model's adherence to the provided CoT, low fidelity scores do not necessarily imply a lack of faithfulness, especially when the model demonstrates the ability to reject misleading information. Further research on CoT faithfulness and the development of more sophisticated evaluation metrics could contribute to the advancement of interpretable and reliable language models.

\subsection{Challenges}
An explanation is considered as faithful if it causes the model to make the same decision as the original input~\citep{li2022faithfulness}. 
In this context, the challenge faced by explainable prompting (e.g., CoT prompt) lies in two aspects: (1) directing language models to generate explanations that are genuinely representative of the models' internal decision-making processes, and (2) preventing language models from depending on potentially biased CoT templates.

Regarding the first challenge, our case study has revealed that relatively small language models may generate answers that do not align with the provided CoT rationales. Therefore, these rationales do not accurately represent the decision-making process within these models. Some efforts have been made to bolster the CoT capabilities of smaller language models by implementing instruction tuning with CoT rationales~\citep{kim2023cot,ho2022large}. 
%These methods can help improve the explanation faithfulness of CoT for small language models, thereby addressing this issue to some extent. 
Nevertheless, it remains a challenging problem of how to ensure the generated explanations (i.e., ``what the model says'') are faithful to the internal mechanism (i.e., ``what the model thinks'') of language models.
Regarding the second challenge, recent research shows that explanations in the CoT can be heavily influenced by the introduction of biasing prompt templates into model input~\citep{turpin2024language}.
%This is because existing CoT requires carefully designed templates to prompt language models to produce explanations.
If incorrect or biased information is encoded in such templates, the generated explanations could be misleading. Recently, \cite{wang2024chain} propose a novel decoding strategy to implement CoT with prompting, which could mitigate this issue. However, how to effectively help language models get rid of the template reliance still remains underexplored.

%\section{LLM Enhancement via Knowledge-Augmented Prompting}\label{sec:prompt_knowledge}
\section{Effective LLM Inference via Knowledge-Augmented Prompting}\label{sec:prompt_knowledge}
Enhancing models with external knowledge can significantly improve the control and interpretability of decision-making processes. While LLMs acquire extensive knowledge through pre-training on web-scale data, this knowledge is embedded implicitly within the model parameters, making it challenging to explain or control how this knowledge is utilized during inference. 
%Additionally, LLMs may not always encompass the unique knowledge specific to certain domains, nor keep pace with the constantly evolving information in the world.
To address these limitations, this section discusses Retrieval-Augmented Generation (RAG) for the explicit integration of external knowledge into LLMs, aiming to yield more interpretable predictions.

\subsection{Retrieval-Augmented Generation with Knowledge}
%By fetching relevant information from external databases or the internet, RAG ensures that LLM outputs are accurate and up-to-date. It addresses LLMs' limitation of relying on fixed and potentially outdated knowledge bases. 
Generally, RAG operates in two steps: (1) \textit{Retrieval}: It locates and fetches pertinent information from an external source based on the user's query; (2) \textit{Generation}: It incorporates this information into the model's generated response. Given an input query $x$ and the desired output $y$, the objective function of RAG can be formulated as~\citep{guu2020retrieval}:
%\begin{equation}
$
   \max_{\phi, \theta}\, \log p(y|x) = \max_{\phi, \theta}\, \log \sum_{z\in \mathcal{K}} p_{\phi}(y|x, z) \cdot p_{\theta}(z|x), 
$
%\end{equation}
where $z$ stands for the external knowledge retrieved from a knowledge base $\mathcal{K}$. Thus, the target distribution is jointly modeled by a knowledge retriever $p_{\theta}(z|x)$ and an answer reasoning module $p_{\phi}(y|x, z)$.
The knowledge $z$ serves as a latent variable. An RAG model is trained to optimize the parameters, so that it learns to retrieve relevant knowledge $z$ and to produce correct answers $y$ based on $z$ and $x$. As LLMs possess stronger text comprehension and reasoning abilities, they can directly serve as the reasoning module $p_{\phi}$ without further training. In this case, RAG can be treated as a data-centric problem:
\begin{equation}
    \max_{z\in \mathcal{K}}\, \log p(y|z, x) = \max_{z\in \mathcal{K}}\, \frac{p(z | x, y)}{p(z | x)} p(y|x) ,
\end{equation}
where the goal is to find appropriate knowledge that supports the desired output. 
The \textit{interpretability} of RAG-based models comes from the information in $z$: (1) $z$ usually elucidates or supplements the task-specific information in $x$; (2) $z$ could explain the generation of output $y$.
Unlike other models that estimate $p(y|x)$ in an end-to-end manner, where the decision process is not comprehensible, the RAG process provides justification or rationale $z$ that supports the result.

% Existing Retrieval-Augmented Generation (RAG) approaches can be categorized based on when they integrate external knowledge into the model's workflow.
% The first category incorporates external knowledge at the \textit{inference} stage. For instance,  \cite{karpukhin2020dense} employ dense vectors to identify related documents or text passages, enhancing the data retrieval step of RAG. Similarly, \cite{lewis2020retrieval} refine the data retrieval process to ensure only the most pertinent information influences the model's output.
% The second category integrates external knowledge during the model \textit{tuning} stage. Some representative approaches include~\cite{guu2020retrieval, borgeaud2022improving, nakano2021webgpt}. Generally, these methods embed a retrieval mechanism into the model's training phase, enabling the model to utilize external data more efficiently from the outset.

\subsection{Enhancing Decision-Making Control with Explicit Knowledge}
%The incorporation of explicit external knowledge through RAG enhances the precision and controllability of decision-making in LLMs. This method leverages real-time information from external databases to produce responses that are not only accurate but also tailored to the specific requirements of each query. Below, we explore the mechanisms by which RAG achieves a more controllable and directed content generation process, with references to key papers that have contributed to these advancements.

\subsubsection{Reducing Hallucinations in Response}
``Hallucination'' in the context of LLMs refers to instances where these models generate information that, while coherent and contextually appropriate, is not based on factual accuracy or real-world evidence~\citep{huang2023survey}. This issue can lead to the production of misleading or entirely fabricated content, posing a significant challenge to the reliability and trustworthiness of LLMs' outputs. 
RAG offers a powerful solution to mitigate the problem of hallucinations in LLMs. By actively incorporating up-to-date, verified external knowledge at the point of generating responses, RAG ensures that the information produced by the model is anchored in reality. This process significantly enhances the factual basis of the model's outputs, thereby reducing the occurrence of hallucinations. \cite{shuster2021retrieval} applies neural-retrieval-in-the-loop architectures to knowledge-grounded dialogue, which significantly reduces factual inaccuracies in chatbots, as confirmed by human evaluations. \cite{siriwardhana2023improving} introduces RAG-end2end, which joint trains retriever and generator components together. Their method demonstrates notable performance improvements across specialized domains like healthcare and news while reducing knowledge hallucination.

\subsubsection{Dynamic Responses to Knowledge Updating} 
RAG empowers LLMs by incorporating the most up-to-date information, keeping their decision-making aligned with the latest developments. This feature is especially vital in fast-evolving fields such as medicine and technology, where the need for timely and accurate information is paramount~\citep{meng2022mass}. For example, research by \citep{izacard2020leveraging} demonstrates significant enhancements in output relevance and accuracy through real-time information retrieval. Similarly, \cite{han2023improving} suggest using retrieved factual data to correct and update the knowledge within pre-trained LLMs efficiently. Additionally, \cite{wang2023retrieval} introduce a method for integrating newly retrieved knowledge from a multilingual database directly into the model prompts, facilitating updates in a multilingual context.

\subsubsection{Domain-specific Customization}
RAG enhances LLMs by incorporating knowledge from specialized sources, enabling the creation of models tailored to specific domains. Research by \cite{guu2020retrieval} illustrates how integrating databases specific to certain fields into the retrieval process can empower models to deliver expert-level responses, boosting their effectiveness in both professional and academic contexts. \cite{shi2023mededit} have applied this concept in the medical domain with MedEdit, utilizing an in-context learning strategy to merge relevant medical knowledge into query prompts for more accurate medical advice. Moreover, recent research finds that LLMs struggle to capture specific knowledge that is not widely discussed in the pre-training data. Specifically,  \cite{mallen2023not} observe that LLMs often fail to learn long-tail factual knowledge with relatively low popularity, finding that simply increasing model size does not significantly enhance the recall of such information. However, they note that retrieval-augmented LLMs surpass much larger models in accuracy.
%, particularly for questions on well-known subjects, suggesting that this method can effectively bridge knowledge gaps.
Similarly, \cite{kandpal2023large} highlights LLMs' challenges with acquiring rare knowledge and proposes that retrieval augmentation offers a viable solution, minimizing reliance on extensive pre-training for capturing nuanced, less common information.

\subsection{Challenges}
We discuss the challenges in RAG that are relevant to its explainability aspects: (1) In the retrieval stage $p_{\theta}(z|x)$, does the retrieved information $z$ always elucidate the task-specific information contained in the input $x$? (2) In the generation stage $p_{\phi}(y|x, z)$, does $z$ effectively serve as an explanation for the generation of output $y$?
%Please note that our goal is not to exhaustively discuss all the limitations of RAG in this paper as RAG itself is a broad topic in NLP research. For a more detailed examination of the broader limitations of RAG, we direct readers to other reviews~\citep{gao2023retrieval}.

\subsubsection{Retrieval Accuracy Bottlenecks}
Existent RAG methods typically rely on similarity search to pinpoint relevant information~\citep{lewis2020retrieval, gao2023retrieval}, which represents a substantial improvement over basic keyword searches~\citep{robertson2009probabilistic}. However, these methods may struggle with complex queries that demand deeper comprehension and nuanced reasoning. 
The recent ``lost-in-the-middle'' phenomenon~\citep{liu2024lost} has revealed that an ineffective retrieval can result in the accumulation of extraneous or conflicting information, negatively affecting the generation quality.
%To address this challenge, recent RAG approaches have integrated adaptive learning processes~\citep{asai2023self}. This advancement enables the retrieval system to refine their performance over time through feedback, adapting to evolving language use and information updates, ensuring their responses remain relevant and accurate. Nonetheless, e
Efficiently handling intricate and multi-hop questions remains a significant challenge, highlighting the need for ongoing research to enhance the capabilities of RAG systems.

\subsubsection{Controllable Generation Bottlenecks}
In-context learning stands out as the premier method for incorporating external knowledge to boost the capabilities of LLMs such as GPT-4~\citep{asai2023self, gao2023retrieval}. Despite its effectiveness, there's no surefire way to ensure that these models consistently leverage the provided external knowledge within the prompts for their decision-making processes~\citep{wu2024faithful,yoran2023making, petroni2020context,li2022large}. 
%In practice, to achieve thorough coverage, commonly used dense retrieval usually returns a large volume of content, including both relevant and redundant information to the input question. Unfortunately, redundant information in the model prompt raises the computational cost and can mislead LLMs to generate incorrect answers. 
%Recent research shows the retrieved information with redundancy can degrade the question-answering task performance~\citep{yoran2023making, petroni2020context,li2022large}. Some recent work proposes to fine-tune the LLM to improve resilience to noise and reduce hallucinations. However, such approach still cannot prevent oversized retrieval information decrease the system interpretability~\citep{yoran2023making,xu2023retrieval}.
The challenge of optimizing the use of external explanations to achieve more precise and controlled decision-making in LLMs is an ongoing issue that has yet to be fully addressed.

\section{Training Data Augmentation with Explanation}\label{sec:data}
This section explores the generation of synthetic data from explanations using LLMs, a technique poised to enhance various machine learning tasks. 
In machine learning, limited data availability often leads to data imbalance and scarcity, which in turn constrains model performance in many domains. A viable solution is data augmentation, where LLMs, with advanced generative capabilities, can be utilized~\citep{whitehouse2023llm,tan2024large}. Nevertheless, several challenges exist for effective text augmentation. First, for utility, the generated samples should exhibit diversity compared to the original data. Second, these samples should exhibit useful patterns relevant to downstream tasks.
To this end, explanation tools could guide augmentations by providing supplemental context and rationales~\citep{carton2021learn}. 

Explanations can be particularly beneficial in two scenarios. 
In the \textit{first scenario}, explanation tools are used to to identify existing deficiencies, which effectively guides the data augmentation. The augmented data include samples and related label annotations. The \textit{second scenario} employs LLMs to directly produce explanatory texts, such as rationales, as supplementary information to enrich the dataset.

\subsection{Label-based Data Augmentation}

For learning tasks, the training dataset includes a large volume of data points, defined as $\mathcal{D}=\{(x_1, y_1), \cdots, (x_n, y_n)\}$. We categorize learning tasks into classification tasks and generation tasks. 

In traditional classification tasks, each data point $x_i$ has a label $y_i$ representing a class. Models are trained to predict the labels for new data points. However, these models are prone to make predictions with spurious correlations also known as shortcuts~\citep{geirhos2020shortcut}. Thus, model's predictions are predominantly based on shortcut features, and the underlying mechanisms of the model are not interpretable from a human perspective. Post-hoc explanation techniques play a crucial role in detecting undesirable correlations between input and predictions~\citep{liu2018adversarial,liu2021adversarial}. For example, \citet{du2021towards} adopt Integrated Gradient (IG) to attribute a model's predictions to its input features, showing that the model tends to treat functional words, numbers, and negation words as shortcuts and rely on them for prediction in natural language understanding tasks. 

LLMs can be applied to resolve data set issues by augmenting various data. For example, explanatory information such as counterfactuals has been incorporated to improve model robustness~\citep{wang2021robustness} and out-of-distribution performance~\citep{sachdeva2023catfood}. LLMs can also synthesize rare examples, which helps models to generalize better on unseen data~\citep{xu2023contrastive}. Besides, another work utilizes attribution-based interpretability tool to identify vulnerable words and leverage LLMs to synthesize natural adversarial examples to enhance model safety~\citep{wang2023generating}. Similarly, LLMs are also helpful in mitigating biases such as fairness issues in models.~\citet{he2023targeted} claims that it automatically identifies underrepresented subgroups, and use LLMs to augment theese subgroups while avoids hurting other groups.

In text generation tasks, each data point consists of a prompt $x$ and a response $y$. The data used for model training is usually text including both prompt and response. Generative models are designed to predict next token based on existing input, which can be formulated as $y_m : \arg \max p_{\theta}(y_m|x, y_{(i\ :\ m-1)})$. To make models generate desired output, well-aligned models are indispensable. Recently, a few work employ LLMs to augment data and enhance model alignment. For example, Wang et al.~\citep{wang2024data} identifies the weakness of existing datasets and instructs LLMs to generate desired data to improve model safety. Another work use attribution-based method to identify key information from privacy-relevant data and replace the privacy information with synthetic information, which provides a way to avoid privacy leakage~\citep{zeng2024mitigating}.

\subsection{Rationale-based Data Enrichment}
LLMs have been leveraged to directly generate natural language explanations as augmented data. This strategy makes use of LLMs' chain-of-thought reasoning abilities~\citep{NEURIPS2022_8bb0d291}. The training datasets contain data with annotated information, denotes as $\mathcal{D} = \{(x_1, r_1, y_1),\ \cdots, (x_m, r_m, y_m)\}$. Each data point includes not only prompt $x$ and response $y$, but reasoning process or annotated information defined as $r$.
As state-of-art LLMs are excellent in interpreting context, many studies have utilized them to provide annotations and enrich data. 

One line of research uses generated explanations to assist and guide model training. ~\citet{li2022explanations} introduce LLM explanations to facilitate the training of smaller models, which improves their reasoning capabilities and acquires explanation generation abilities. Hsieh et al~\citep{hsieh-etal-2023-distilling} integrate LLM rationales as additional supervision to guide the training of smaller models. Experiments have shown that this approach not only requires fewer training data but also outperforms traditional fine-tuning and distillation methods. Another line of study employ annotated information to boost models' performance on specific tasks. For example, LLM explanations have also been utilized to generate explanations for sentiment labels of aspects in sentences to mitigate spurious correlations in aspect-based sentiment analysis tasks~\citep{wang2023reducing}. Besides, Code translation generation tasks incorporate explanations as an intermediate step, improving model performance by 12\% on average~\citep{tang2023explain}. The result shows that explanations are particularly useful in zero-shot settings. Wei et al.~\citep{wei2024instructrag} instruct LLMs to explain the retrieval process from documents to produce rationales. Then they can be used to train models and further improve models ability on retrieval-augmented generation tasks. Similarly, Lupidi et al.~\citep{lupidi2024source2synth} prompt LLMs to generate data explanations from real-world sources such as articles, which are then used to fine-tune models and improve their performance on multi-hop reasoning and information retrieval.

\subsection{Challenges}
\subsubsection{Computational Overhead}
Conventional post-hoc explanations, built on well-trained models, are often resource-intensive tasks. The first scenario mentioned above leverages interpretability techniques to accurately diagnose dataset issues. This process typically requires multiple rounds of model training and applying interpretability methods to develop fair and robust models. Consequently, the crafting process can be both time- and energy-consuming. 
% Given these challenges, exploring the development of data-centric evaluation metrics is crucial. These metrics can offer a more efficient way to assess data issues, bypassing traditional, cumbersome explanation methods. By focusing on these data-centric measurements, data issues can be diagnosed and fixed before training. The number of training rounds needed is then significantly reduced. This shift not only streamlines model development but also helps reduce computational overhead, making the whole process more practical and efficient.

\subsubsection{Data Quality and Volume}
Despite their advanced capabilities, LLMs still have limitations when dealing with highly specialized or niche contexts. For example, one of the most prominent issues is ``hallucination'', where models generate plausible but incorrect or misleading responses. This could adversely affect the quality of augmented data, potentially introducing more biases to which LLMs are also vulnerable. 
% Another challenge is controlling the relevance of LLM-generated content. That is, the explanations or data points may seem reasonable but often lack factual accuracy or nuances specific to a domain. Currently, we lack robust metrics to effectively measure the quality and relevance of these generated data relative to the original tasks. 
Determining the precise amount of data required is also challenging, often leading to new dataset imbalances. Managing the quantity of LLM-generated data is an immense challenge, as augmented data can introduce other biases~\citep{zhao2021lirex}. This stems from LLMs' limited ability to accurately control the quantity and distribution of generated data. 
% Moreover, crafting effective prompts is more of an art than a science, adding uncertainty around generated data quality. 
Together, these factors underscore the complexities and challenges in fully harnessing LLMs' potential for data augmentation and related tasks.

\section{Generating User-Friendly Explanation with LLMs}\label{sec:user-friendly}
Previous sections mainly focused on quantitative explanations with LLM via numerical values. For example, sample-based explanation discussed in Section \ref{sec:sample} aims to assign each training sample an influence score (see Eqs.\ref{eq:influence_func}-\ref{eq:emb}) that measures the confidence that we can use that training sample to explain the prediction of a test sample. However, using numerical values for explanations is not intuitive, which can be difficult to understand by practitioners with little domain knowledge~\citep{latif2024fine,lee2023multimodality,li2020generate,chen2024selfie}. User-friendly explanations, on the contrary, aim to generate human-understandable explanations, e.g., natural language-based descriptions, regarding why a model makes certain predictions or what role a neuron plays in the network, such that the explanations can be well-understood by both researchers and practitioners.

% Given an explainee $e$, which can be a data sample $(x_{i}, y_{i})$, a neuron $\theta_{i}$ from a pretrained model $f_{\theta}$, or a prediction result $\hat{y}$ based on the input $x$, generating user-friendly explanation aims to map the explainee $e$ to a sequence of natural language tokens as the explanation for the explainee $e$, such that the generated explanations can be easily comprehended by human beings.

% \subsection{User-friendly Data Explanation with LLMs}

% Data explanation refers to the process of translating difficult materials (e.g., program codes, long documents) into concise and straightforward language so that they are easy to understand by humans. 
% Language models have long been used to generate explanations for textual data~\citep{dai2019deeper}. Since LLMs are trained on corpora composed of codes, math, and papers, they can be leveraged to explain data beyond pure textual content. For example, \cite{chen2021evaluating} demonstrate that pretrained GPT models possess the ability to understand and generate codes, where explanatory comments are generated simultaneously that facilitate the understanding of programmers. In addition, \cite{welleck2022naturalprover} propose to explain math theorems by providing detailed derivations, so that the theorems are easier to understand. Recently, LLMs have also been used to elucidate academic papers \citep{castillo2022chat}, making difficult content to be easily understood by individuals with little domain knowledge.

\subsection{Explaining Small Models with LLMs}

Recently, there has been growing interest in leveraging LLMs to generate free-text explanations for small models. For example, to explain black-box text classifiers, \cite{bhattacharjee2023llms} propose a prompting-based strategy to identify keywords $K=\{k_{1}, k_{2},...,k_{n}\}$ in the input texts $x$ with pretrained LLMs that are informative for the label $y$, and ask LLMs to substitute them with another set of keywords $K'=\{k'_{1}, k'_{2},...,k'_{n}\}$, such that changed text $x'$ changes the label prediction to $y'$. They view the textual mapping rule ``if we change $K$ into $K'$ in $x$, then $y$ will be classified as $y'$” as the counterfactual explanation for the model. In addition, to explain the neuron of a pretrained language model (e.g., GPT2), \citet{bills2023language} propose to summarize the neuron activation patterns into \textit{textual phrases} with a larger language model (e.g., GPT4), where the neuron activation patterns are expressed as a sequence of (token, attribution score) pairs. To verify the identified patterns, they generate activation patterns according to the phrases via the same LLM and compare their similarity with the true activation patterns of the neuron, where the phrases with high scores are considered more confident to serve as the explanation for the neuron. 

The explaining ability of LLMs is not necessarily limited to text models. For example, \cite{zhao2023automated} propose using pretrained vision-language models to generate explanations for a neuron $\theta_{i}$ of an image classification model. Specifically, for each class $y = y_{c}$, they first find regions in images with label $y_{c}$ that have maximum activation of the neuron $\theta_{i}$ as the surrogate explainees for $\theta_{i}$, and prompt LLMs such as ChatGPT to generate candidate explanations (words, short phrases) for the class label $y_{c}$. Then, they use the pretrained vision-language model CLIP~\citep{radford2021learning} to match the candidate explanations with the surrogate explainees as the explanations for the neuron $\theta_{i}$. Recently, LLMs have also found applications in explaining recommender systems \cite{zhu2023causal}. Specifically, \cite{yang2023large} found that LLMs can well interpret the latent space of sequential recommendation model after alignment, whereas \cite{lei2023recexplainer} propose to align user tokens of LLMs with the learned user embeddings of small recommendation model to generate explanations of user preferences encoded in the embeddings. Recently, ~\cite{schwettmann2024find} propose a unified framework to explain all models where inputs and outputs can be converted to textual strings. Specifically, the explainer LLM is used as an agent to interact with the explainee model by iteratively creating inputs and observing outputs from the model, where the textual explanations are generated by viewing all the interactions as the context.

\subsection{Self-Explanation of LLMs}

Due to the black-box nature of LLMs, it is promising to generate user-friendly explanations for the LLMs themselves, such that their operational mechanics and the predictions can be understood by human experts. Based on whether the LLM needs to be retrained to generate explanations for themselves, the self-explanation of LLM can be categorized into two classes: \textit{fine-tuning based} approach and \textit{in-context based} approach, which will be introduced in the following parts.

\textbf{Fine-tuning based approaches.}
Given sufficient exemplar explanations on the labels of the training data (e.g., in recommendation datasets such as the Amazon Review datasets \citep{he2017translation} or the Yelp dataset \citep{zhou2020s3}, users have provided explanations on why they have purchased certain items, which can be viewed as explanations for the ratings), LLMs can learn to generate explanations for their predictions as an \textit{auxiliary task} through supervised learning. One exemplar method is P5 \citep{geng2022recommendation}, which fine-tunes the pre-trained language model T5 \citep{raffel2020exploring} on both the rating and explanation data to generate an explanation alongside the recommendations. Recently, several works have improved upon P5 \citep{cui2022m6,zhu2023collaborative}, which fine-tunes different LLMs such as GPT2, LLaMA, Vicuna, etc., and propose different prompt learning strategies \citep{li2023personalized} with generating explanation as the auxiliary task.
With explanations introduced as additional supervision signals to fine-tune pretrained LLMs for recommendations, the performance can be improved with good explainability.

\textbf{In-context based approaches.} In many applications, there is often a lack of sufficient exemplar explanations. However, the unique capability of modern LLMs to reason and provide answers through human-like prompts introduces the potential for in-context based explanations. Here, explanations for predictions are crafted solely based on the information within the prompt. A leading approach in this domain is the Chain-of-Thoughts (CoT) prompting~\citep{wei2022chain}, which provides few-shot examples (with or without explanations) in the prompt and asks the LLM to generate answers after reasoning step-by-step, where the intermediate reasoning steps that provide more context for generating the final answer can be viewed as explanations.
However, CoT generates reasoning first and then based on which generates predictions, where the reasoning steps can influence prediction results \citep{lyu2023faithful}. If explanations are generated after the prediction, since the explanation is conditioned on the predicted label, it can provide a more faithful post-hoc explanation of why the model makes certain decisions \citep{lanham2023measuring}. 
The application of in-context based self-explanation of LLMs is broad. For example, \cite{huang2023can} explore generating zero-shot self-explanation of sentiment analysis with LLMs by directly asking them to generate explanations alongside the predictions. In addition, \cite{huang2023chain} propose a chain-of-explanation strategy that aims to explain how LLMs can detect hate speech from the textual input. \cite{lu2022learn} find that CoT can generate well-supported explanations for question answering with scientific knowledge.

\subsection{Challenges}

\subsubsection{Usability v.s. Reliability}
Many existing methods rely on prompts to generate user-friendly explanations, which are not as reliable as numerical methods with good theoretical foundations. \cite{ye2022unreliability} find that the explanations by CoT may not be factually grounded in the inputs. Therefore, they believe that these explanations are more suitable as post-hoc explanations regarding why the LLM makes certain predictions. 
However, the validity of viewing CoT explanations as post-hoc justifications has been questioned by recent findings~\cite{turpin2024language}, which uses biased datasets (e.g., the few-shot examples in the prompt always answer ``A'' for multiple choice questions) to show that the generated explanations may be plausible, but systematically unfaithful to represent the true reasoning process of the LLMs. Therefore, there's a growing need for more theoretical scrutiny of user-friendly explanations to ensure faithfulness and credibility.

\subsubsection{Constrained Application Scenarios}

Currently, the utilization of LLMs to explain smaller black-box models is mainly limited to those that deal with data with rich textual information \citep{bhattacharjee2023llms,lei2023recexplainer}. Although \cite{zhao2023explainability} propose a strategy to explain image classifiers, the ability to match candidate textual explanations with image patterns still relies on the pretrained vision-language model CLIP. Therefore, there is a compelling need to development more versatile strategies for explaining models across a wider range of fields. This endeavor could depend on the fundamental research on combining LLM with other domain-specific tasks, such as the development of Graph-Language Models that are applicable to unseen graphs in a zero-shot manner.

\section{Interpretable AI System Design with LLMs}\label{sec:architecture}
An intriguing but challenging problem in XAI is creating model architectures or even AI systems that are inherently interpretable~\citep{rudin2019stop}, where different model components represent clear and comprehensible concepts or functionalities that are easily distinguishable from one another.
Machine learning models such as support vector machines~\citep{hearst1998support} and tree-based models~\citep{song2015decision} were classical techniques for achieving model interpretability. 
In the deep learning era, typical research areas in this context include concept-bottleneck models~\citep{koh2020concept,yuksekgonul2022post}, disentangled representation learning~\citep{denton2017unsupervised,higgins2016beta}, and network dissection~\citep{bau2017network,bau2018gan}.
Nevertheless, under the traditional deep learning setting, the usability of these techniques remains limited because of two major challenges. First, it is difficult to define the spectrum of concepts or functionalities the model is expected to capture. Second, the efficacy of interpretable models often falls short compared to black-box models, thereby constraining their practical utility.

Large foundation models, such as large language models (LLMs) and vision language models (VLMs), provide opportunities to bridge the gap. 
By leveraging the common-sense knowledge embedded within them, foundation models can \textit{design interpretable architectures} by providing cues that encourage creating and using the features or procedures within AI workflows. This is different from traditional deep learning pipelines, where the deep models automatically discover the features during the training process, which may not end up with model components with clear meanings. 
Furthermore, LLMs can decompose complex tasks into simpler and collaborative sub-tasks, enhancing both the system's interpretability and its overall performance.

\subsection{Designing Interpretable Network Architectures with LLMs}

Representative methods for developing interpretable deep architectures include Generalized Additive Models (GAMs)~\citep{agarwal2021neural,zhuang2021interpretable,lou2012intelligible} and Concept Bottleneck Models (CBMs)~\citep{koh2020concept,yuksekgonul2022post,shang2024incremental}. These models map inputs into a human-understandable latent space, and then apply a linear transformation from this space to the target label. 
For example, to build a classifier that diagnoses arthritis, we can let the model identify features such as ``bone spurs'' and ``sclerosis'', and then use these interpretable features for the final decision.
However, these approaches often require the involvement of experts to define the latent space, which can limit the learning capabilities of deep models.
Some work tries to automate the discovery of semantic concepts during model training, such as by requiring independence between concepts~\citep{higgins2016beta, yu2020learning} or clustering data~\citep{ghorbani2019towards}, but they lack direct control over the outcomes and does not ensure the clarity of the concepts. One promising strategy is to utilize LLMs to provide comprehensible concept candidates. \cite{menon2022visual} use human language as an internal representation for visual recognition, and create an interpretable concept bottleneck for downstream tasks.  
By basing the decision on those comprehensible concepts, the model architecture itself is provided with better transparency. Similarly, a recent approach \textit{Labo}~\citep{yang2023language} constructs high-performance CBMs without manual concept annotations. This method controls the concept selection in bottlenecks by generating candidates from the LLMs, which contain significant world knowledge~\citep{petroni2019language} that can be explored by prompting a string prefix. Human studies further indicate that those LLM-sourced bottlenecks are much factual and groundable, maintaining great inherent interpretability for model designs. Besides the concept-based models, another promising strategy is to employ LLMs to enhance the conventional architectures that are inherently interpretable, such as GAMs and Decision Trees (DTs). \cite{singh2023augmenting} leverages the knowledge captured in LLMs to enhance GAMs and DTs, where LLMs are only involved during the augmented model training instead of the inference process. 
For GAMs training, LLMs can provide decoupled embeddings for enhancement. For DTs training, LLMs are able to help generate improved features for splitting. The LLM-augmented GAMs and DTs enable full transparency, where only the summing coefficients and input key phrases are required for interpretation. With the extra information from LLMs, augmented GAMs and DTs are capable of achieving better generalization performance compared with non-augmented ones.

\subsection{Designing Interpretable AI Workflows with LLM Agents}
Traditional deep models are usually designed in an end-to-end manner. The internal workflows are not quite understandable to general users. By utilizing common-sense world knowledge, LLMs can break down complex problems into smaller ones and organize the workflows among them, leading to a more interpretable design of AI systems~\citep{feng2023knowledge}. A recent example of interpretable AI workflow design comes from ~\cite{shen2024hugginggpt}, where an LLM-powered agent leverages ChatGPT to integrate various off-the-shelf AI models (e.g., from Hugging Face~\citep{jain2022hugging}) to handle different downstream application tasks. In order to handle complicated tasks in a transparent workflow, LLMs play a pivotal role in coordinating external models with language mediums to harness their powers. By planning the target task, selecting candidate models, executing decomposed subtasks and summarizing responses, LLMs can help disassemble tasks based on user requests, and assign appropriate models to the tasks based on the model descriptions. Similarly, to transparentize the workflow, \cite{liu2023controlllm} introduces a task decomposer to analyze the user prompts and break it down into a number of subtasks for solving using LLMs. Each subtask is well managed and attributed with \textit{description}, \textit{domain}, \textit{inputs}, and \textit{outputs}. In this way, the AI systems are then capable of handling intricate user prompts with a step-by-step understandable workflow. Under the prompting paradigm, \cite{khot2022decomposed} also employs LLMs to solve complex tasks by decomposition. Drawing inspiration from software libraries where the workflows are trackable, the decomposer and shared subtasks are designed in a modular manner. One step further, \cite{wang2024describe} introduces an interactive planning approach for complex tasks, which enhances the error correction on initial LLM-generated plans by integrating plan execution descriptions and providing self-explanation of the feedback. Such an interactive nature enables better workflow transparency in long-term planning and multi-step reasoning task scenarios.

\subsection{Challenges} 
\subsubsection{Planning Feasibility in Complicated Scenarios} 
Despite the task planning capability of LLMs, it is still challenging to apply to certain scenarios in real-world applications due to feasibility issues. One typical scenario is the few-shot planning cases~\citep{guo2023multimodal}, where acquiring large datasets for training is either impractical or cost-prohibitive, thus making feasible planning about unseen cases from sparse exemplars extremely challenging. To better assist the interpretable designs, LLM planning needs to generalize well without extensive supervision and is expected to have the ability to integrate information from prior experiences as well as knowledge. Besides, another important scenario lies in the dynamic planning settings~\citep{dagan2023dynamic}, in which LLMs integrate feedback from the environment iteratively, letting the agent take thinking steps or augment its context with a reasoning trace. Dynamic scenarios urgently and frequently involve high computational costs resulting from the iterated invocations of LLMs, and still face challenges in dealing with the limits of the context window and recovering from hallucinations on planning.

\subsubsection{Assistance Reliability with Knowledge Gaps}  
LLMs exhibit remarkable proficiency in encapsulating real-world knowledge within their parameters, but they resort to hallucinations and biases with high confidence when certain knowledge is missing or unreliable. 
% Although a growing number of techniques has been proposed, such as retrieval augmentation~\citep{guu2020retrieval}, searching integration~\citep{nakano2021webgpt} and multi-LLM collaboration~\citep{feng2023knowledge}, to expand LLM knowledge, such discrepancy in knowledge may perpetually exist owing to the continuously evolving character of human understanding~\citep{ji2023survey}. 
As a result, a crucial research challenge keeps rising, \textit{i.e.}, how to effectively detect and mitigate the LLM knowledge gaps from humans when employing LLMs for designs. We will need further research on evaluating and developing robust LLM mechanisms to address the knowledge-gapping problems, with the goal of helping improve LLM reliability, reducing hallucinations and mitigating biases. Furthermore, the intersections between the knowledge gaps and the safety aspects are also a great challenge to be solved, which may pose some security concerns, especially when using LLMs for downstream models or workflow designs.

\section{Emulating Humans with LLMs for XAI}\label{sec:mimic_humans}
This section discusses how LLMs can play the role of humans in serving XAI development. Building explainable models requires two main steps where humans are in the loop: (1) collecting a dataset with human-annotated rationales to train the models; (2) collecting human feedback on the quality of explanations produced by the models for evaluation. The significant cost and time required for human involvement raise the main challenge in scaling up this procedure. LLMs emerge as a promising solution to this challenge, thanks to their capability to emulate human reasoning and produce responses that closely resemble human-generated content.
In the following, we introduce the methods that demonstrate LLMs' ability to generate human-like annotations and feedback for XAI.

\subsection{Emulating Human Annotators for Training Explainable Models} 
Incorporating human-understandable rationales into model development has shown its effectiveness in enhancing both the transparency and performance of the system for various NLP tasks, such as question answering~\citep{li2018vqae,wu2020improving}, sentiment analysis~\citep{du2019learning,antognini2021rationalization}, and common sense reasoning~\citep{rajani2019explain,camburu2021rationale}. 
We use the term \textit{rationales} to describe supportive evidence that justifies the connection between inputs and outputs~\citep{gurrapu2023rationalization}. 
Traditionally, the rationales are collected by leveraging human annotations~\citep{camburu2018snli,wang2019learning} or applying expert-designed rules~\citep{alhindi2018your,li2018vqae}, resulting in expensive costs or limited quality. 
Recently, researchers in \textbf{automatic annotation}~\citep{ding2022gpt,belal2023leveraging,gilardi2023chatgpt} have begun to explore the potential of leveraging advanced LLMs to emulate human annotators in annotating the target labels of task-specific examples.  
These studies found that advanced LLMs show comparable annotation qualities against average crowd human annotators on most tasks with a lower cost, pointing out the scalability of using machine-emulated annotators. 
Inspired by these works, some studies~\citep{huang2023chatgpt,huang2023chain,pang2024iterative,yang2024weak,wu2024thinking,zhang2024learn} attempt to leverage advanced LLMs to collect rationales by applying the \textbf{chain-of-thought} prompting. 
Specifically, researchers provide several input-rationale-output demonstrations within the input text to prompt the LLMs to generate rationale and output for an unlabeled input instance. 
The quality of such annotated rationales largely relies on the in-context learning capabilities of LLMs, leading to uncontrollable annotation quality on uncommon tasks. 
Other scholars~\citep{yao2023beyond,chen2023zara,luo2023xal} propose a \textbf{human-in-the-loop} LLM-based annotation framework based on the \textbf{active-learning} architecture.    
This framework initially collects a small seed dataset with human-annotated rationales and labels. This seed dataset is used to train an explainable classifier for this downstream task. Then, each unlabeled sample is passed through the trained explainable classifier. This is followed by a selection strategy that chooses representative samples according to metrics such as explanation plausibility, prediction uncertainty, and sample diversity. Finally, LLMs are leveraged to annotate the rationales and labels of these selected unlabeled samples. This procedure could be repeated multiple times, and the trained explainable classifier from the latest time is the final output of this framework. 
Compared with other methods, this approach balances the annotation quality and the cost budget in developing explainable models by using LLM-emulated annotators.

\subsection{Emulating Human Feedback for Evaluating Explainable Models}
The explanations generated by the explainable models could be classified into two categories: extractive and abstractive~\citep{gurrapu2023rationalization}.
Extractive explanations derive directly from the input data, exemplified by attribution-based methods that emphasize specific segments of the input text. In contrast, abstractive explanations are generated in a free-form text manner, such as chain-of-thought (CoT) responses~\citep{wei2022chain}, offering a more nuanced interpretation. The quality of extractive explanations is typically assessed through their \textbf{agreement with human-annotated rationales}~\citep{deyoung2020eraser}, such as accuracy, recall, and precision. 
However, evaluating abstractive explanations presents a significant challenge, as it is impractical to exhaustively evaluate all reasonable abstractive results comprehensively.
To automatically assess abstractive explanations, early studies first collect some free-text rationales, and then compute \textbf{the cosine similarity between embeddings of explanations and rationales}~\citep{cheng2023explainable,li2023ucepic}. 
A higher similarity between the abstraction explanation and the annotated rationales indicates a more transparent model.
Recently, some researchers have followed the \textbf{LLM-as-a-judge} framework to ask LLM to judge the model explanations without referring to any human-annotated rationales~\citep{miao2023selfcheck,bills2023language,zheng2023judging,saha2025learning,wei2024systematic}, emphasizing the potential of emulating human feedback with advanced LLMs.

\subsection{Challenges} 
\subsubsection{Uncontrollable Credibility of Emulation} 
While LLMs could assist in rationale collection and explanation evaluation, their provided results can be biased and may not always match human annotators, due to hallucinated responses in their unfamiliar domains~\citep{ji2023survey} and that LLMs are still immature~\citep{harding2023ai}. 
% This issue leads to unreliable annotations or feedback, as LLMs confidently generate factually incorrect conclusions.
The quality of data gathered from this process is compromised, impacting the development of XAI systems. 
To improve the quality of annotations and feedback, future research could focus on incorporating hallucination detection~\citep{dhuliawala2023chain} and retrieval augmented generation~\citep{ren2023investigating} techniques. These methods could enhance the reliability of LLM outputs, making them more comparable to human-generated content in the context of XAI development. 

\subsubsection{Ethical Considerations in LLM Annotation}
When LLM annotators keep human annotators away from subjective scenarios, such as hate speech detection~\citep{huang2023chatgpt}, LLMs also have a chance to inject unethical opinions into their annotated datasets. 
Although most advanced LLMs are fine-tuned to align with human values~\citep{ouyang2022training}, such as being helpful, honest, and harmless, many studies have shown that this protection mechanism can be jailbroken~\citep{wei2023jailbroken,zou2023universal}, causing the model to produce values-violating answers. 
Ensuring LLM annotators follow ethical guidelines is worth further exploration.

\section{Conclusion}
In this paper, we discuss a crucial yet frequently underappreciated aspect of Explainable AI (XAI) -- \textit{usability}.
To this end, we present 10 strategies for advancing Usable XAI within the LLM paradigm, including (1) leveraging explanations to reciprocally enhance LLMs and general AI systems, and (2) enriching XAI approaches by integrating LLM capabilities. 
Unlocking the potential of XAI's usability can help address various challenges in LLM such as human alignment. 
We also provide case studies to several critical topics, aiming to provide resources for interested developers.
We further discuss open challenges at the end of each strategy, suggesting directions for future work in this evolving area.

\bibliographystyle{ACM-Reference-Format}
\bibliography{sample-base}

\end{document}